\documentclass[letterpaper, 10 pt, conference]{ieeeconf}  

\IEEEoverridecommandlockouts                              

\usepackage{float}
\usepackage{subfigure}
\usepackage{multirow}
\usepackage{graphicx} 
\usepackage{amsmath} 
\usepackage{hyperref}

\usepackage{verbatim}

\usepackage{algorithm}
\usepackage{algorithmicx}
\usepackage{algpseudocode}
\floatname{algorithm}{Algorithm}

\title{\LARGE \bf
	Area Graph: Generation of Topological Maps using the Voronoi Diagram 
}

\author{Jiawei Hou, Yijun Yuan and S\"oren Schwertfeger 
	\thanks{The authors are with the School of Information Science and Technology, ShanghaiTech University
			Shanghai 201210, China, 
	and also with the University of Chinese Academy of Sciences,
			Beijing 100049, China 
			{\tt\small [houjw, yuanwj, soerensch]@shanghaitech.edu.cn}}
}

\begin{document}

	\maketitle
	\thispagestyle{empty}
	\pagestyle{empty}

	\begin{abstract}

    Representing a scanned map of the real environment as a topological structure is an important research topic in robotics. 
    Since topological representations of maps save a huge amount of map storage space and online computing time, they are widely used in fields such as path planning, map matching, and semantic mapping.
    
	We use a topological map representation, the Area Graph, in which the vertices represent areas and edges represent passages. The Area Graph is developed from a pruned Voronoi Graph, the Topology Graph. We also employ a simple room detection algorithm to compensate the fact that the Voronoi Graph gets unstable in open areas. We claim that our area segmentation method is superior to state-of-the-art approaches in complex indoor environments and support this claim with a number of experiments.

	\end{abstract}

\section{INTRODUCTION}
	Robotics has seen tremendous developments in recent years. There are more and more mobile autonomous robots that are active in bigger and more complex areas for long times. This poses challenges for the storage and computation with traditionally used 2D grid maps for big areas, the canonical output of most Simultaneous Localization and Mapping (SLAM) algorithms.
	The obvious solution to this problem is to use topological map representations, a method already used by car navigation systems. 
	
	
	In our work, we present a novel method to extract a topological representation, the Area Graph, from a 2D grid map. We extract areas in the environment based on the Topology Graph presented in \cite{schwertfeger2012robotic}, \cite{schwertfeger2016map} and \cite{Schwertfeger2016PathMatching}. From those areas, a graph where the vertices represent the areas and the edges represent the common boundaries between two areas, i.e \emph{passages}, is created as the Area Graph. 

We believe that our representation is more useful than classical topological representations, because it is representing areas instead of places, which allows a more intuitive usage of topological maps. When humans think of a place, they very seldom mean a singular point, but an area of a certain shape and size, e.g. a city, a house or a room. An area graph representation also makes it very clear to which area specific coordinates belong, while this is not so easy for classical point and line-based topological representations.

Topological representations with areas as well as segmentation has been used for quite a while. Already 1998 Thrun \cite{Thrun1998Learning} proposed to use critical points along Voronoi Graphs for segmentation. Other, often similar segmentation
algorithms have been presented since: \cite{Kai2008Coordinated} \cite{bormann2016room} \cite{Malcolm2017MAORISICRA}. In contrast to the mentions approaches, our algorithm is directly anchored on the topology of the Voronoi Graph, so it is separating areas only on junction points. This is well motivated later with the maze experiment in Figure \ref{fig:maze}. 
The room detection we also employ is mainly a method to avoid over-segmentation. 

	The main contributions of our paper are the novel method to generate the Area Graph from 2D grid maps and the experiments comparing our method to the state-of-the-art 
	to show the excellent segmentation ability in complex indoor environments of our method.

	The paper is structured as follows. In Section~\ref{sec:generation_of_areagraph}, 
	the Topology Graph presented in \cite{schwertfeger2012robotic} and \cite{schwertfeger2016map} is briefly described
	and the approach for the Area Graph generation is explained in detail.
	Since the Area Graph segments the environment into areas, we present, in Section~\ref{sec:comparison_segmentation}, related works on map segmentation. We subsequently compare the results of our algorithm with state of art segmentation algorithms. 
	Conclusions are drawn in Section~\ref{sec:conclusions}.
	

	
\section{Generation of the Area Graph}
\label{sec:generation_of_areagraph}
	In this section, two 2D preprocessing steps are shortly presented and  a brief overview of the steps for generating a Topology Graph \cite{schwertfeger2016map} is recalled.
	
	Then, a detailed description of the algorithm which generates areas for the Area Graph is presented. The method is twofold: first, areas are generated from the Topology Graph, then areas in the same room are merged together.

\subsection{Preprocessing}
\begin{figure*}[tpb]
		\subfigure[Map before preprocessing.]{
			\label{fig:input_removal}
    	\includegraphics[width=0.32\linewidth]{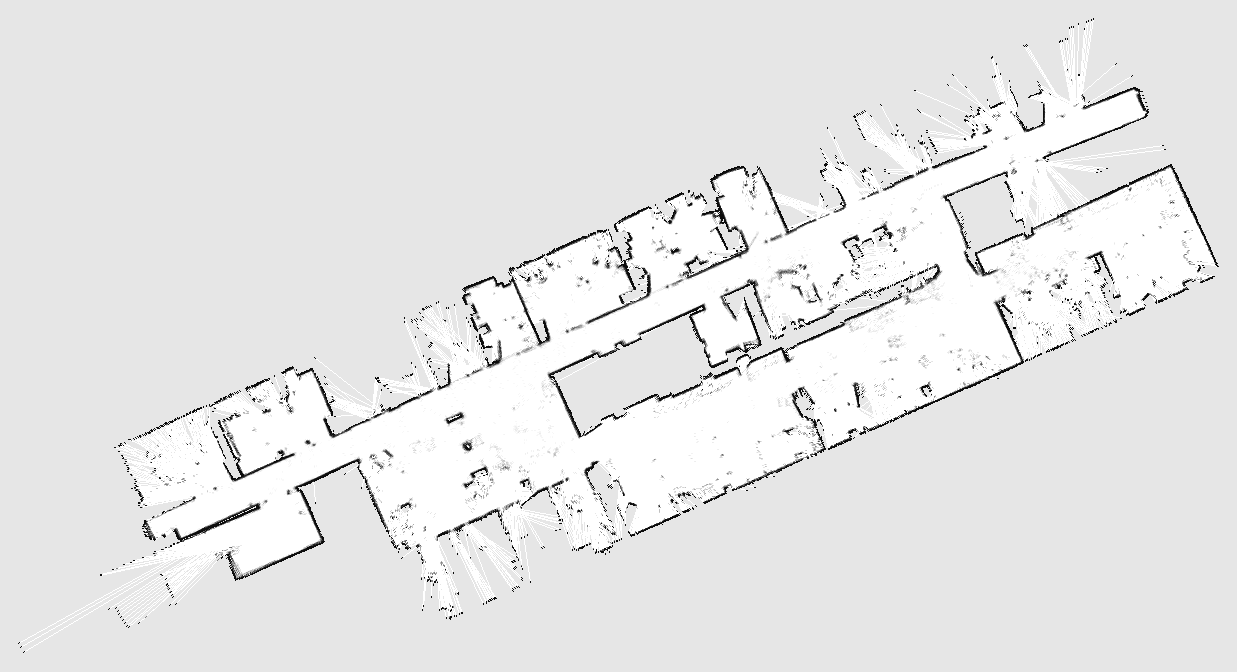}}
		\subfigure[Map after outliers removal.]{
			\label{fig:outliers_removal}
    	\includegraphics[width=0.32\linewidth]{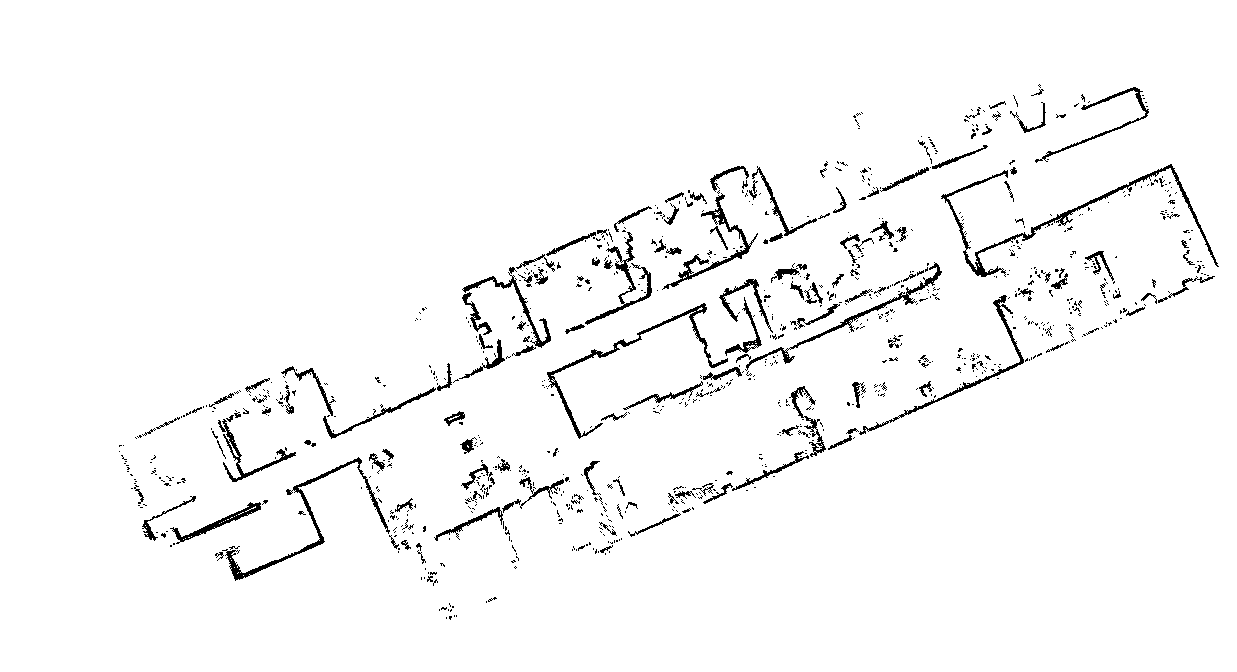}}
		\subfigure[Map after furniture removal.]{
			\label{fig:furniture_removal}
    	\includegraphics[width=0.32\linewidth]{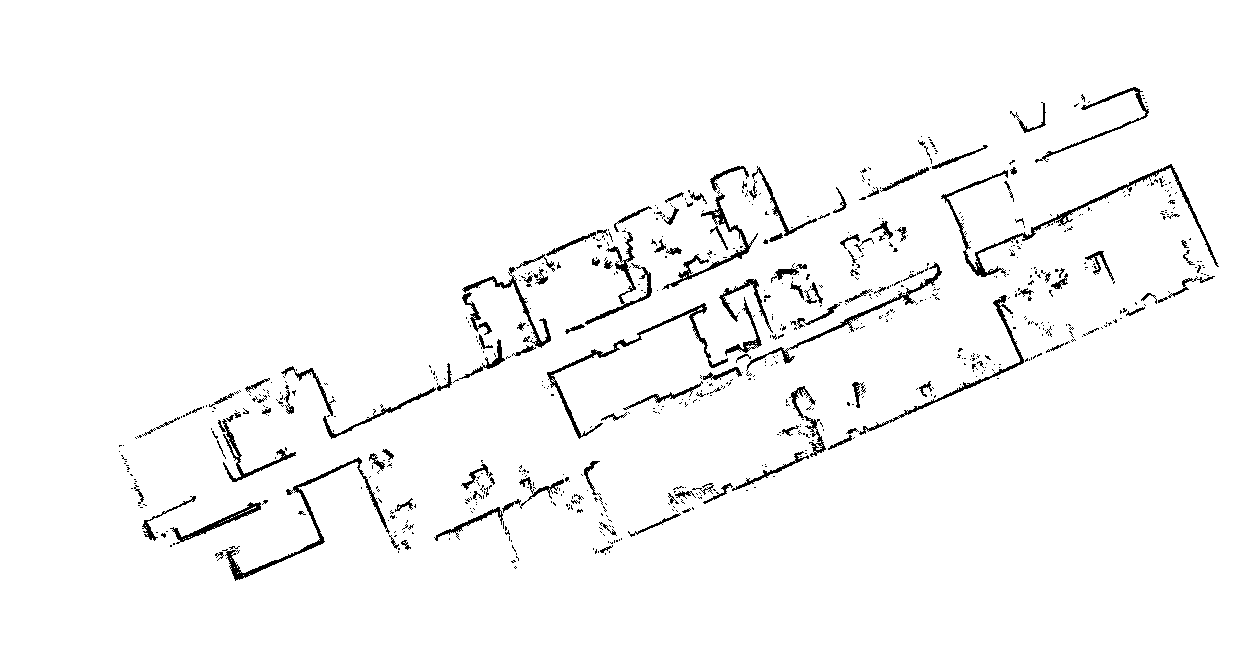}}
    	\caption{2D Grid Map Preprocessing.}
	\end{figure*}

    The main cause of over-segmentation usually are noise and furniture. To reduce the effect of the noise, we use the \emph{remove\_outliers} function by CGAL to initially remove the outliers from the input image. Then we detect $\alpha$-shape with $\alpha$ computed from $W_{robot}$ to detect small pieces of points which are largely furniture, where $W_{robot}$ is the width of robot. Fig. \ref{fig:outliers_removal} and \ref{fig:furniture_removal} show the maps after outliers removal and furniture removal, respectively.

\subsection{Topology Graph Generation}
	\label{sec:topo_generation}
	
		
	

	The Topology Graph \cite{schwertfeger2016map} is developed from a Voronoid Diagram (VD) of the 2D grid map. 
	
	Because the VD extends to infinity outside of the map, 
	the boundary of the map needs to be found to remove the unwanted data. 
	For that purpose the alpha shape algorithm \cite{Edelsbrunner1983On} is used. 
	The algorithm uses the CGAL 2D Alpha Shapes \cite{cgal:d-as2-17b} to generate alpha shapes, and the biggest alpha shape is regarded as the boundary. All the vertices and edges outside such boundary are filtered out.
	
    The VD from the 2D point map has edges going between two close obstacles point, i.e. points at walls, so all edges whose distances to occupied cells that are smaller than a threshold are deleted from the graph.
	This processing leaves only dead-ends and junctions as vertices of the graph. Dead-ends below certain lengths are then removed (several times) to create the Topology Graph.
	
\begin{figure*}[tb]
			\label{fig:lab2}
			\centering
			\subfigure[]{
			\label{fig:topoGraph}
				\includegraphics[width=0.32\linewidth]{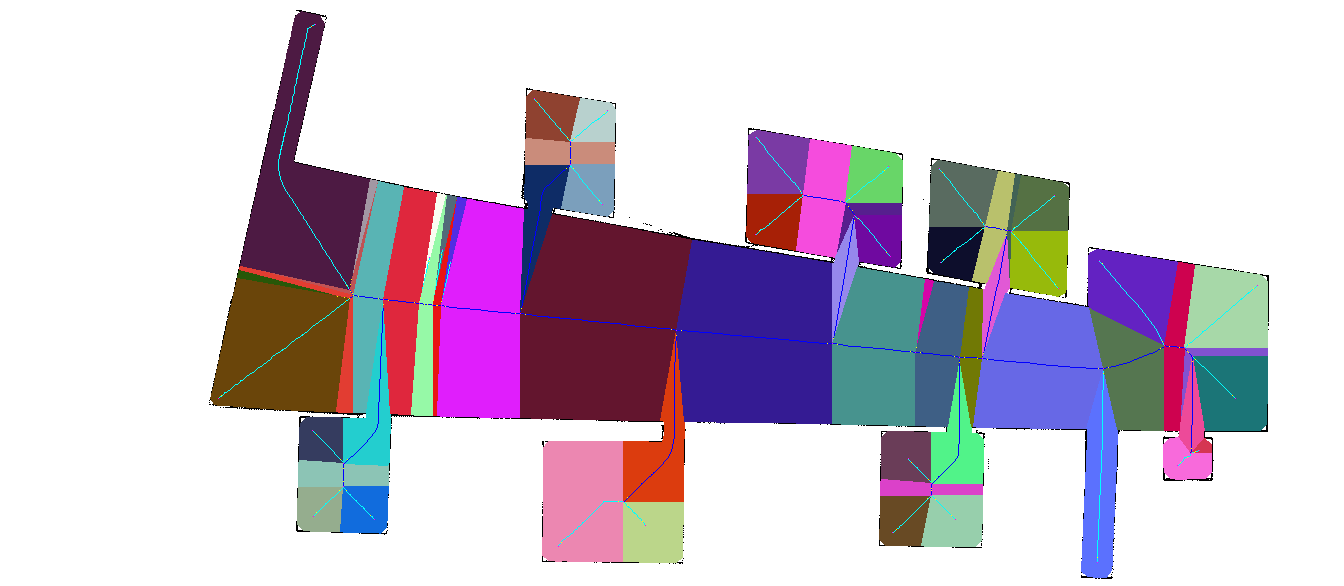}}
			\subfigure[]{
				\label{fig:alphaLab}
				\includegraphics[width=0.32\linewidth]{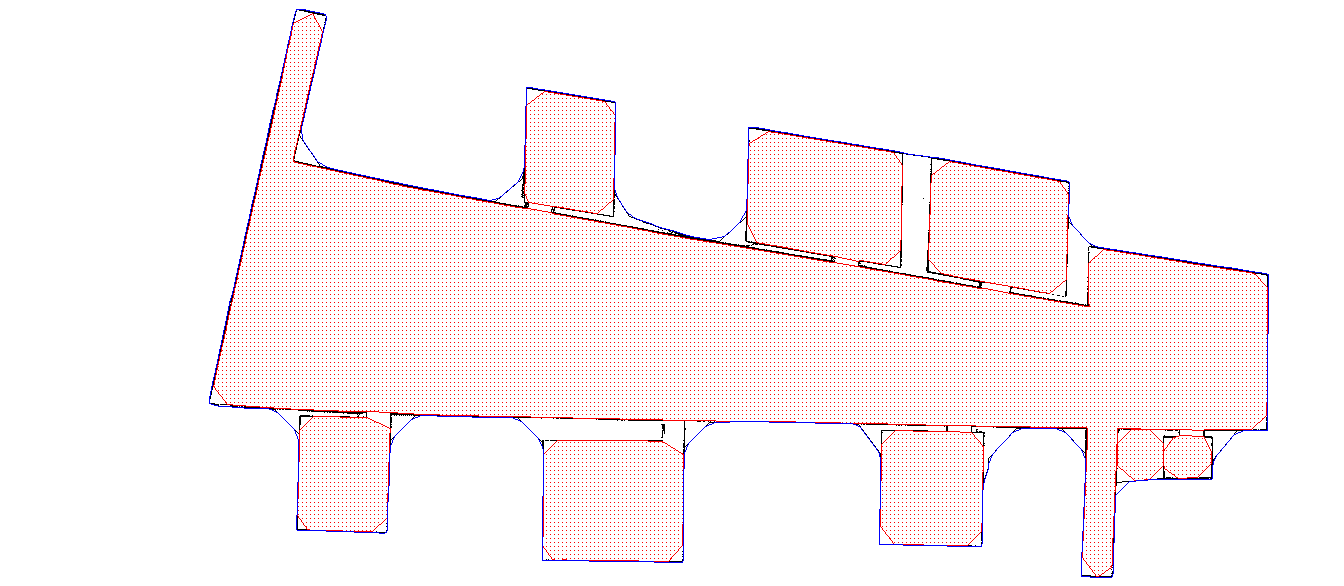}}
			\subfigure[]{
				\label{fig:dectRoomLab}
				\includegraphics[width=0.32\linewidth]{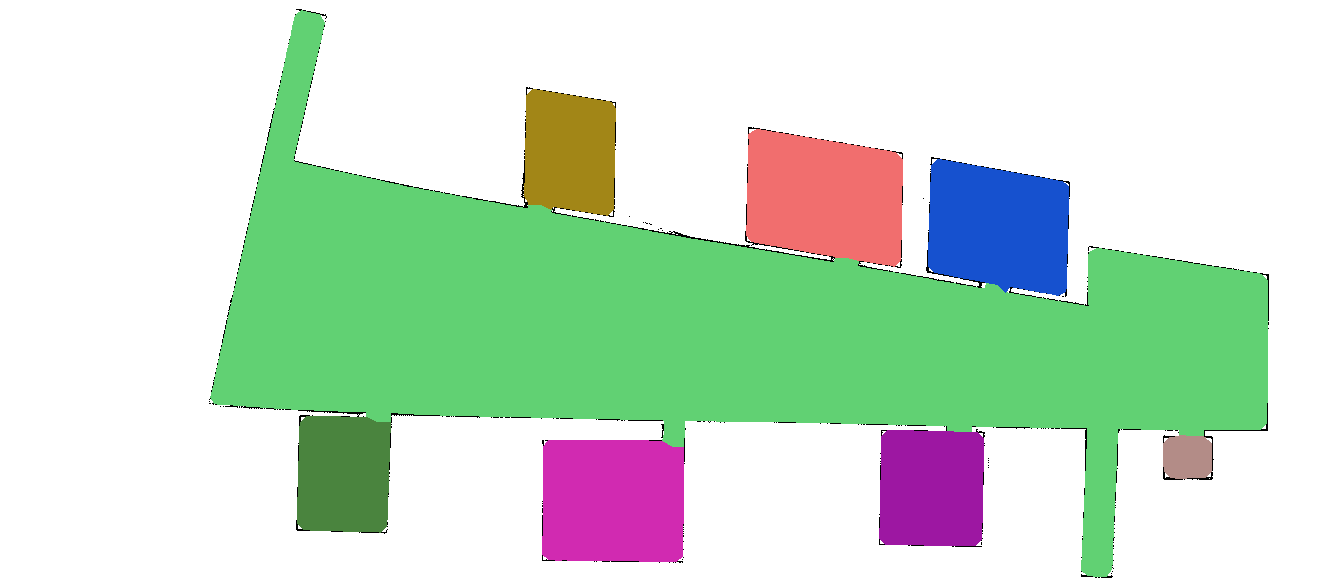}}
			\caption{
			(a) The Topology Graph obtained by pruning the Voronoi diagram using the method in \cite{schwertfeger2016map}. Dead-end edges are shown as light-blue lines and edges connecting non-dead-end vertices (junctions) are shown as dark blue skeleton.  The polygons attached on the Topology Graph are areas have not be merged.
			(b) Alpha shape detection to get the boundary (blue line) and rooms (red patterns) from the grid map.
			(c) Areas within the same alpha shapes have been merged.
}
		\end{figure*}

	Since only the graph of reachable areas is useful, the connected graph with the largest sum of length is kept and all vertices and edges not belonging to the biggest connected graph are removed. Finally, vertices that are too close are merged.
	Fig. \ref{fig:topoGraph} shows a Topology Graph constructed from a grid map.

	\subsection{Generating Areas from a Topology Graph}

    \begin{algorithm}[tb]
    \caption{Generate the Area Graph\label{alg:main}}
    \begin{algorithmic}[1]
    
        \Require 2D Grid Map $map$
        \Ensure Area Graph $G_{A}=(V_A,E_A)$

        \State $map \gets$ \Call{OutliersRemoval}{$map, noise\_percent$}
        
        \State $map \gets$ \Call{AlphaShapeRemoval}{$map, \alpha_{robot}$}
        
        \State $VD \gets $ \Call{createVoriGraph}{$map$}
        
        \State $ biggestPoly, alphaShapes \gets $ \Call{performAlphaShape }{$ map, \alpha_m$}
            
            
        \State $VD \gets $ \Call{removeOutside}{$VD, biggestPoly$}
        
        \State $repeat\_cnt = 3$
        \While{$repeat\_cnt--$}
        \State $VD \gets $ \Call{removeDeadends}{$VD$}
        
        \State $G_{A}^0 \gets $ \Call{joinHalfEdgePolys}{$VD$}
        \EndWhile
        
        \State $G_{A}^0 \gets $ \Call{keepBiggestGroup}{$G_{A}^0$}
        
        \State $G_{A}^0 \gets $ \Call{removeRays}{$G_{A}^0$}
        
        \State $G_{A}^1 \gets $ \Call{cutEdges}{$G_{A}^0, alphaShapes, biggestPoly$}
        
        
        \State $G_{A} \gets $ \Call{MergeAreas}{$G_{A}^1, alphaShapes$}
    \end{algorithmic}
    \end{algorithm}



		\begin{figure}[t]
			\begin{center}
				\includegraphics[width=1.0\linewidth]{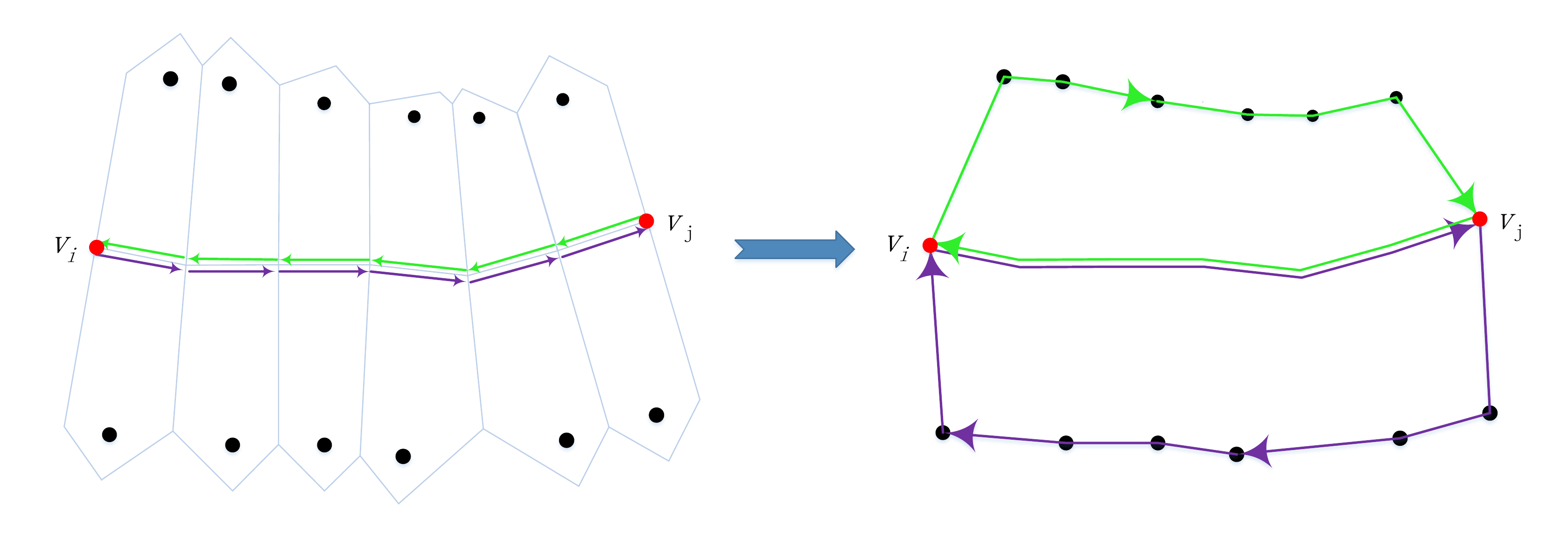}
			\end{center}
			\caption{Left schematic: The green short arrows are the Voronoi edges that make up the halfedge $ h_{ji} $. Correspondingly, the purple short arrows make up the half edge $ h_{ij} $ in posite direction. Each Voronoi edge has a face, which contains exactly one site inside. Right schematic: Connect two halfedges with sites in clockwise order respectively to create two half-polygons for an edge.} 
			\label{fig:faces}
		\end{figure}
		\begin{figure}[t]
		
			\subfigure[Join a polygon of a dead end edge to its next half-polygon, in the case that its next neighbor edge is not a dead-end.]{
				\label{fig:rightdeadend}
				\includegraphics[width=0.95\linewidth]{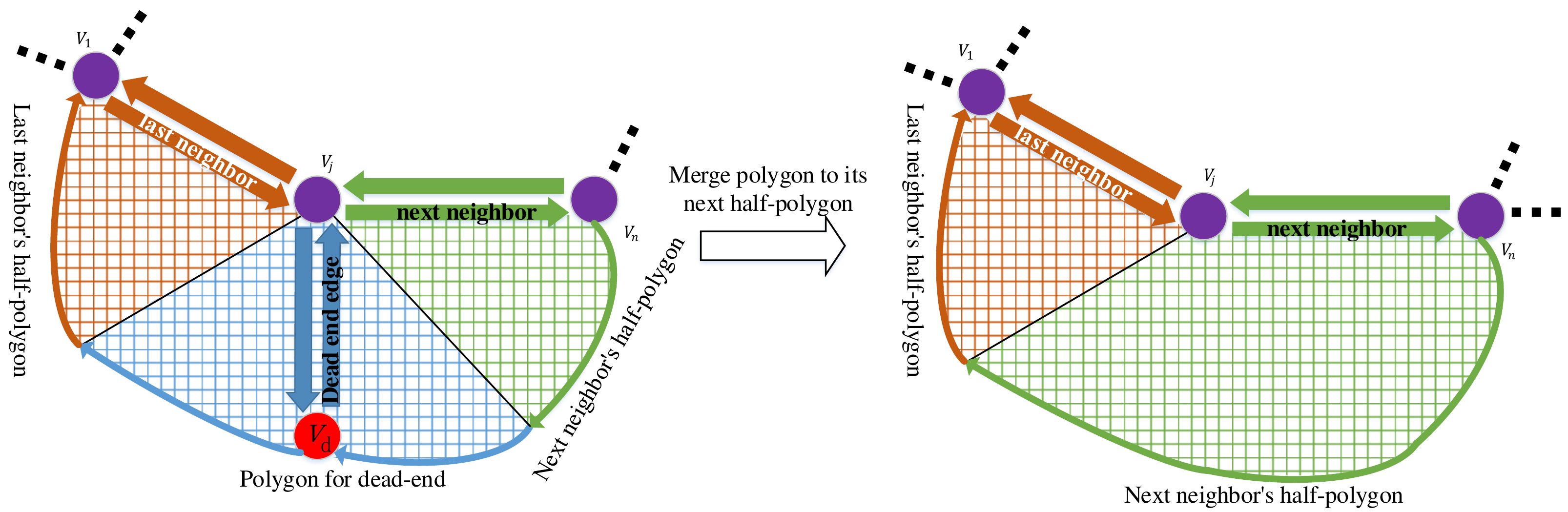}}
			\subfigure[Join a polygon of a dead end edge to its last  half-polygon, because the right neighbor of $h_{dj}$ is a dead-end.]{
				\label{fig:leftdeadend}
				\includegraphics[width=0.95\linewidth]{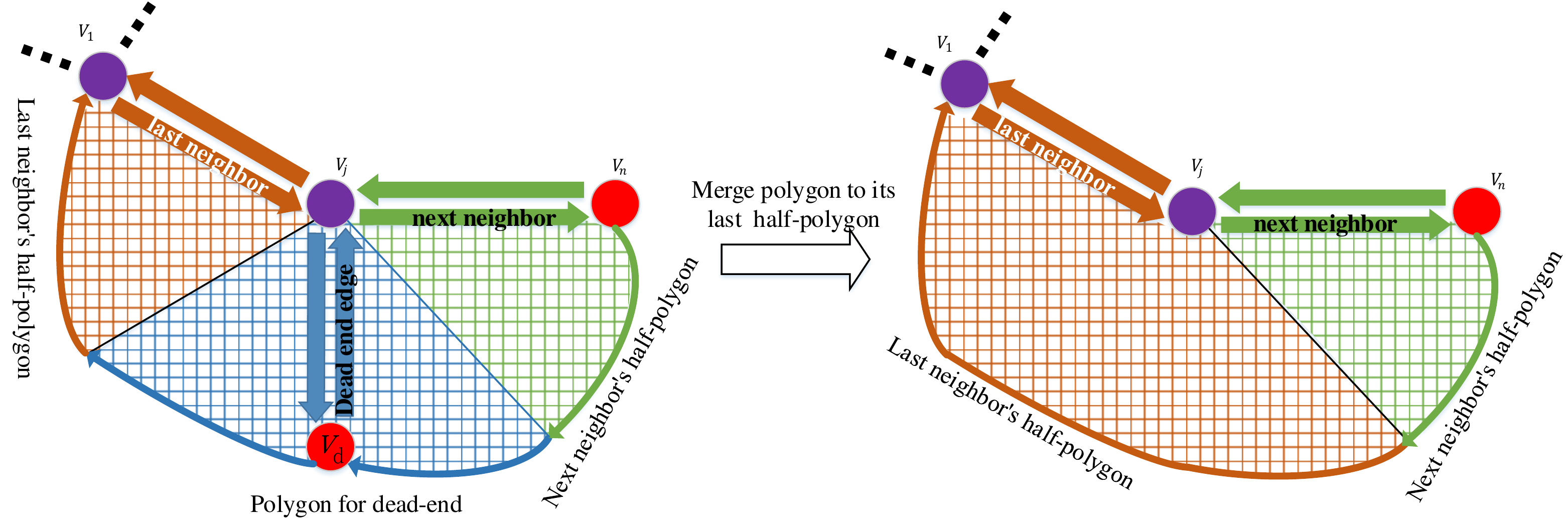}}
			\caption{The two cases of merging polygon of the dead-end edge into its neighbor's half-polygon (here only shows half polygons of $e_{dj}$'s neighbors instead of the whole polygon). Red points represent dead-ends, the edge connected to a dead-end is a dead-end edge. Purple points represent junctions. \label{fig:deadend_face}}
		\end{figure}
		
    \begin{algorithm}[t]
    \caption{Remove a deadend and its polygon\label{alg:deadendremoval}}
    \begin{algorithmic}[2]
        \Function{removeDeadends}{$e_{dj}$}
            \State $h_{dj},h_{jd}\gets half\_edge(e_{dj})$
            \State $hp_{dj} \gets half\_polygon(h_{dj})$ 
            \State $hp_{jd} \gets half\_polygon(h_{jd})$
            \State $h_{lj} \gets last\_halfedge(h_{jd})$
            \State $hp_{lj} \gets half\_polygon(h_{lj})$
            \State $h_{jn} \gets next\_halfedge(h_{dj})$
            \State $hp_{jn} \gets half\_polygon(h_{jn})$
            \If{IsDeadend($h_{jn}$)}
                \State $hp_{lj}.points\gets h_{lj}.points+hp_{dj}.sites+hp_{jd}.sites+hp_{lj}.sites$
            \Else
                \State $hp_{jn}.points \gets hp_{jn}.points+hp_{dj}.sites+hp_{jd}.sites$
            \EndIf
        \EndFunction
    \end{algorithmic}
    \end{algorithm}
    

    The Area Graph generation, as shown in Algorithm \ref{alg:main}, is using the same steps as the Topology Graph generation, 
    but now we keep and merge the information about the areas belonging to each edge of the Voronoi Diagram in every step. An area of our graph thus corresponds to an edge in the Topology Graph, 
    where, for each Topology Graph edge, all the faces of the VD have been joined. 
 
	First, the Voronoi Diagram $VD = (V_{VD}, E_{VD}, F_{VD})$ is generated with the Computational Geometry Algorithms Library (CGAL) \cite{cgal:k-vda2-17b}.  
	Then the part of the graph outside the boundary is removed and edges that are shorter than a user-defined threshold are filtered out.
	
	The modified version of edge skipping \cite{schwertfeger2016map}, as shown in Fig. \ref{fig:faces}, is run to generate the first level Area Graph  $VG^0_A$, which can be seen as the Topology Graph with polygons attached to edges.
	
	\begin{align*}
	    G^0_A =& (V^0_A, E^0_A, P^0_A),\\
	    \text{where } \qquad&\\
	    V^0_A =& \{ v \in V_{VD} | deg(v)=1 \cup deg(v)>2 \\
	           & \cup (deg(v)==2 \cap \exists e \in v | ray(e)) \} ,\\
	    E^0_A =& \{ e_{ij}=e_{ji}=(v_i, v_j)=(h_{ij}, h_{ji}) | v_i, v_j \in V^0_A\} ,\\
	    P^0_A =& \{ p_{ij}=poly(e_{ij})=(hpoly(h_{ij}), hpoly(h_{ji})) \} . 
	\end{align*}

	An edge of the graph $ e_{ij}=(v_i, v_j) $ consists of two twin halfedges $ \{h_{ij}, h_{ji} \}  $ that are in opposite direction.  During the VD generation, the faces of the halfedges are already saved \cite{cgal:k-vda2-17b}. These faces can be utilized to create a polygon for each edge. 
	
	For each halfedge $ h_{ij} $, we build a half-polygon $ hpoly(h_{ij}) $ by connecting the waypoints and the dual sites in the faces in clockwise order. The process is shown in Fig. \ref{fig:faces}. A pair of twin half polygons $ hpoly(h_{ij})$ and $hpoly(h_{ji}) $ are regarded as the polygon attached to the edge $ e_{ij} $, denoted as $ p_{ij}=poly(e_{ij}) \in P^0_A$.

	When the edges are joined, the polygons attached to the edges are merged. We merge the half-polygons of two joined halfedges by connecting the waypoints of the two halfedges and their sites in clockwise order. 
	
	
	We also employ a dead-end removal for short dead-end edges to simplify the graph. 
	When a dead-end edge is removed, the polygon attached to that edge needs to be merged into its neighboring polygons. Otherwise, the areas of the dead-ends would be lost.
	To make it easier to understand, we show the process in Fig. \ref{fig:deadend_face} and Algorithm \ref{alg:deadendremoval}.
	If a  dead-end  edge  $ e_{dj}=\{h_{dj},h_{jd} \} $ needs to be deleted,
	its polygon $ p_{dj}=poly(e_{dj}) $ will be merged into one of its two neighboring half-polygons. 
	Here we label the halfedge $ h_{jn} $ after $ h_{dj} $ in clockwise order as the \emph{next halfedge} of $ h_{dj} $, and the halfedge $ h_{lj} $ before $ h_{jd} $ in clockwise order as the \emph{last halfedge} of $ h_{dj} $. The terms \emph{last half-polygon} and \emph{next half-polygon} are defined in the same way.
	We prioritize merging the dead-end edge's polygon $poly(e_{dj})$ to its next half-polygon $ hpoly(h_{jn}) $. If $ h_{jn} $ is also a dead-end halfedge, then $poly(e_{dj})$ will be merged into its last half-polygon $ hpoly(h_{lj}) $. This strategy helps to avoid the imbalance of the polygon area for dead-end edges.
	Fig. \ref{fig:deadend_face} shows both cases for merging the polygon of a dead-end.

	\subsection{Merging Areas in the Same Room}
	 \label{sec:mer_room}
	The VD becomes unstable in open areas or rooms. That means, that small variations on the walls of rooms with have big impact in the topology of the VD in such open areas. Also, ideally we do not want to segment a room into different areas, it should be just one big area in our graph. So we employ a room detection algorithm to merge areas in a room. We again use the  CGAL 2D Alpha Shapes, this time for detecting the rooms. 
	
	In Fig. \ref{fig:alphaLab}, we show an example of alpha shapes extracted from a map.  
	The open space inside the boundary detected by $ \alpha$-shape is regarded as a \emph{room}. 
	The minimum area of the room that can be detected depends on the parameter $ \alpha $, which is the square of the radius of the largest empty disk that can be put into the detected rooms.
	The larger $ \alpha $, the larger the smallest $\alpha$-shape is, and fewer rooms are detected.
	
	
	The $ \alpha $ value is calculated from a parameter: the width $W$. 
	To detect rooms, the most important point is to ensure that the virtual disk cannot pass through the door. Thus, the radius of the disk cannot be smaller than half the width of the door $ W_d $.
	If we want to segment the whole corridor as one area, the diameter of the virtual disk should be smaller than the width of the narrowest corridor $ W_c $. Therefore, when $ W_d < W_c $, a reasonable choice of $W$ is the one satisfies $ W_d < W < W_c $. Section \ref{subsec:experiment_alpha} analyzes a strategy of choosing $ \alpha $ values.
	
	First, we convert the unit of the chosen width from meters to pixels.
	Then $ \alpha $ can be computed from $ W $ by:  
	\begin{align*}
	    W_{pixel} = \frac{W} {\text{Resolution}} , \quad \alpha = R^2  = (\frac{W_{pixel}}{2})^2 .
	\end{align*}

	After obtaining the $\alpha$-shapes to find rooms, polygons in the same room need to be merged as an area to represent a complete room. For this purpose, we split the polygons crossing the boundary between rooms, and merge the polygons belonging to the same room.

    \begin{algorithm}[t]
    \caption{Cut edges across the alpha shapes}
    \label{alg:cutedges}
    \begin{algorithmic}[2]
        \Function{cutEdges}{$G_{A}^0, alphaShapes, biggestPoly$}
        \Require $G_{A}^0=(V_{A}^0, E_{A}^0, P_{A}^0), \alpha-shape list, biggest \alpha-shape$
        \Ensure $G_{A}^1=(V_{A}^1, E_{A}^1, P_{A}^1)$
        \State $E_A^1 \gets E_A^0$
        \While{True}
            \For{$e \in E_A^1$}
             \For{$s_\alpha \in alphaShapes$}
                \If{$e$ and $s_\alpha$ intersect at point $p$}
                    \State $toCutMap[e] \gets p$
                \EndIf
             \EndFor
            \EndFor
            \If{$toCutMap.isEmpty()$}
                break;
            \EndIf
            \For{$e \in toCutMap$}
                \State $p \gets toCutMap[e].front()$
                \State $e_1, e_2 \gets$ \Call{cutEdgeandPoly}{$e,p$}
                \State $E_{A}^1 \gets e_1, e_2$
                \State $E_{A}^1.remove(e)$
            \EndFor
        \EndWhile    
        \EndFunction
    \end{algorithmic}
    \end{algorithm}
	If an edge crosses the $ \alpha $-shape, i.e. one of its endpoints is inside and the other endpoint is outside the $\alpha$-shape and the edge is not a short dead-end, its polygon will be divided into two at the \emph{passage point}. A passage point is the intersection of the edge and the $\alpha$-shape. The process is shown in Algorithm \ref{alg:cutedges}.
	
	All polygons belonging to the same room are assigned the same $ roomID $. Then we merge the polygons with the same $ roomID $ into one polygon, representing a room. Fig. \ref{fig:dectRoomLab} shows the area after merging polygons.

		
	Through the above steps, an Area Graph can be created. The areas are regarded as vertices of the Area Graph. The neighbors for each area are recorded and the passages between areas are the edges of the Area Graph connecting adjacent areas.
	
\subsection{Evaluation the Influence of $\alpha$ \label{subsec:experiment_alpha}}

\begin{figure}[h]
	\subfigure[$W$-MCC curve]{
		\label{fig:alpha_mcc}	\includegraphics[width=0.99\linewidth]{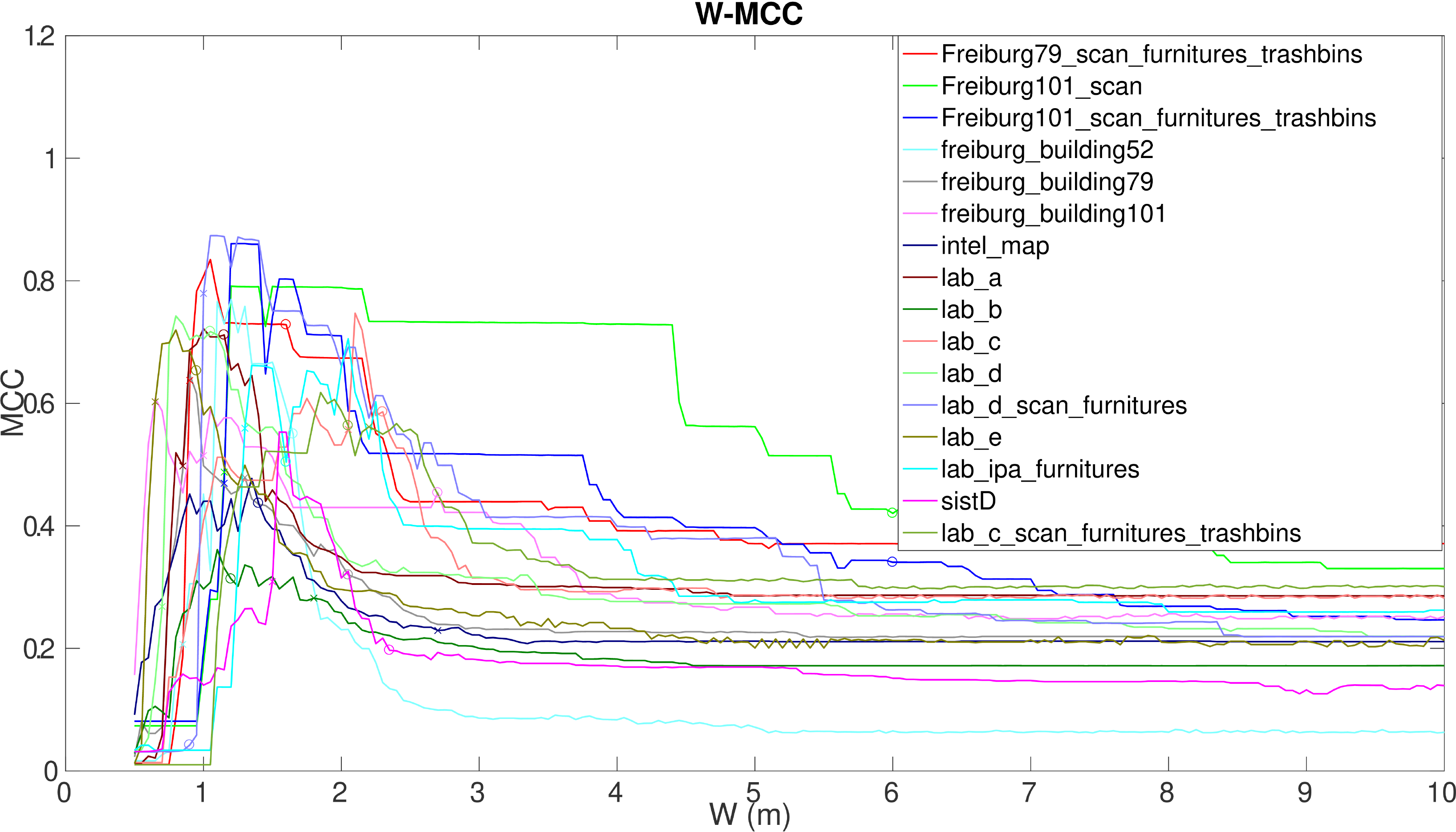}}
	\caption{Evaluation of segmentation quality to $W$, which decides the $\alpha$ in our algorithm. The MCC score at $W=W_d$, the widest door width, and $W=W_c$, the narrowest corridor width, are marked with "x" and "o", respectively, for each curve. \label{fig:exp1}}
\end{figure}

Now follows an experiment to analyze the influence of the $\alpha$ value,  using Matthew's correlation coefficient (MCC) \cite{Malcolm2017MAORISICRA} for segmentation measurement against a human-made ground truth. 
The range of MCC is from -1 to 1, where with the best segmentation MCC=1. 
Because our algorithm is designed for real environment segmentation, the experiments are run on Bormann's sub-dataset, including the non-artificial maps and the furnitured maps, and a map scanned from a part of our building.

We evaluate the segmentation results to $\alpha$ by running the program with the width from $W=0.5 m$ to $W=10 m$, considering a door would not be narrower than $0.5m$ while an indoor corridor is hardly wider than $4m$. The $W-MCC$ curves are shown in Fig. \ref{fig:exp1}. 

\begin{figure*}[tb]
	\subfigure{
		\label{fig:underseg}	\includegraphics[width=0.31\linewidth]{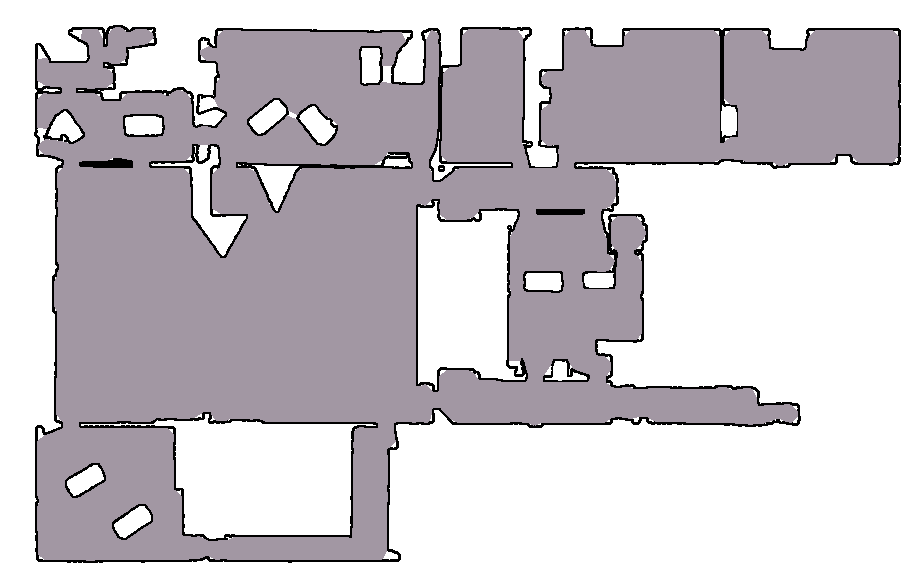}}
	\subfigure{
		\label{fig:goodseg}	\includegraphics[width=0.31\linewidth]{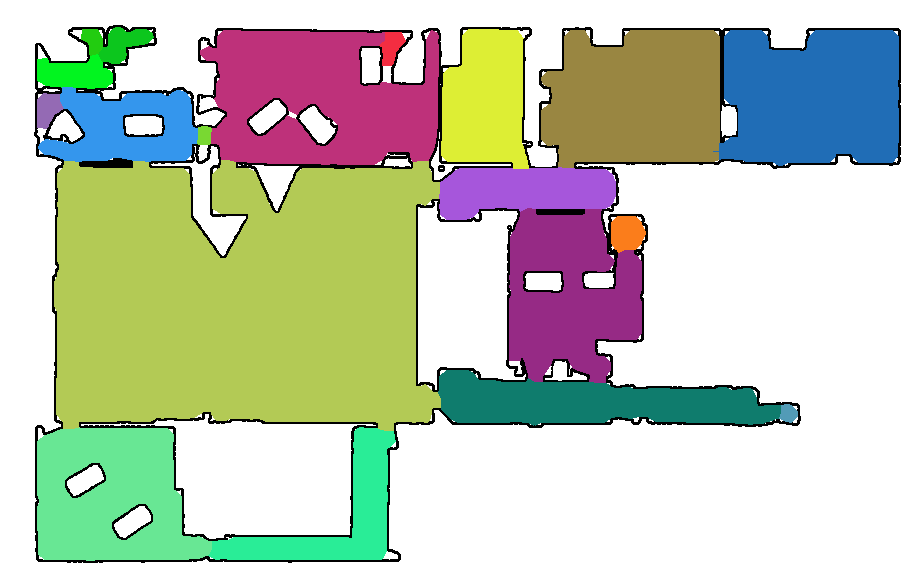}}
	\subfigure{
		\label{fig:overseg}	\includegraphics[width=0.31\linewidth]{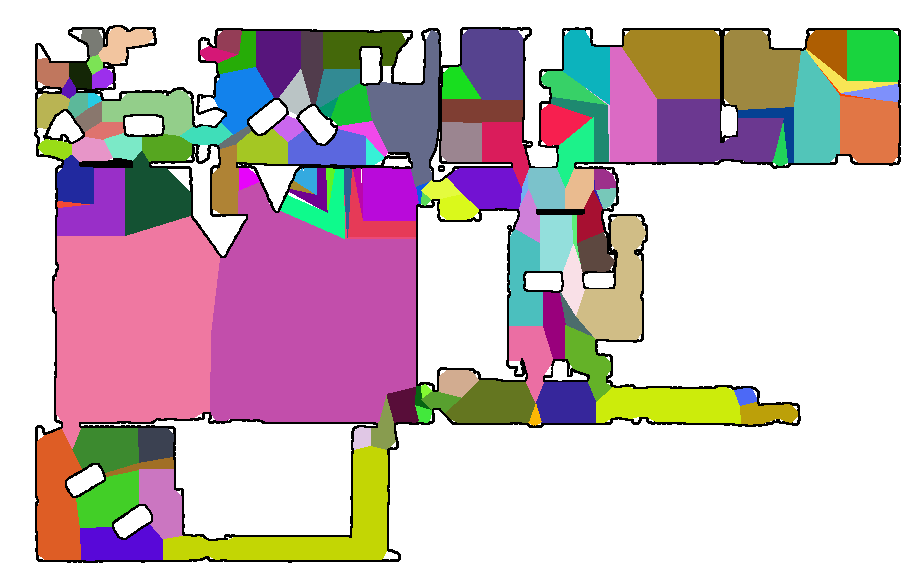}}
	\caption{The segmentation results with different $W$s (0.5m, 0.8m and 8.55m). The MCC score of the three results in Fig. \ref{fig:seg52} are 0.03, 0.74 and 0.23. \label{fig:seg52}} 
\end{figure*}

Fig. \ref{fig:exp1} shows the results of the alpha shape experiment. We notice that for small $W$, the alpha shape goes through the doors, thus one alpha shape will engulf the entire space, thus merging all areas into just one (see Fig. \ref{fig:underseg}): we have under-segmentation. 

On the other extreme, large values for $W$ result in very few to no rooms being detected, leaving the pure Topology Graph-based area algorithm, which suffers from over-segmentation in rooms (see Fig. \ref{fig:overseg}).

Observe Fig. \ref{fig:exp1}, the sweet spot for $W$ and thus $\alpha$ is a size somewhat bigger than the widest door when $ W_d < W_c $, while for the case of $ W_d > W_c $, the peak of the curve happens at the $W$ that is slightly smaller than the $ W_c $. Therefore, we computed the mean of $ W_{best} - W_d $ for the ones with $ W_d < W_c $, which equals to $ 0.1m $, and the mean of $ W_{best} - W_c $ for the ones with $ W_d > W_c $, which equals to $ -0.1m $. Finally, we suggest the strategy to use the $ W = W_d + 0.1m $ for the maps with $ W_d < W_c $, while chose the $ W = W_c - 0.1m $ in the case $ W_d > W_c $.

\section{Comparing with Works on Segmentation}
\label{sec:comparison_segmentation}
	
	The comparison between the Area Graph and other methods in segmentation is shown with a discussion and an experiment.
	
	\subsection{Compare with State-of-the-art Segmentation}
	
	Fundamentally, a basic contribution of the Area Graph is that it represents a map as a set of connected areas, which can be regarded as a segmentation for a map. 
	

	In \cite{Thrun1998Learning}, a Voronoi graph-based segmentation method is shown, which creates a topological representation using region cells that represent rooms or parts of rooms by means of critical points. A \emph{Critical point} is the point lying closer to obstacle points than all its neighboring points on the Voronoi graph. Thus critical points usually locate at narrow passages such as doorways. 
	After that, the Voronoi-based segmentation works \cite{Kai2008Coordinated} with some optimization to select the critical points only at real doors is proposed.
	To segment complete rooms, some heuristics are used to merge small segments \cite{bormann2016room}. 
	 
	Bormann et al. \cite{bormann2016room} introduced various kinds of methods on room segmentation and compared the segmentation results of those methods. 
	And the comparison experiment of \cite{bormann2016room} shows that Voronoi-based segmentation results have the highest degree of approximation of the ground truth. Bormann implemented the Voronoi-based segmentation based on the algorithm from \cite{Thrun1998Learning} and added more heuristics.
	 
	 Mielle et al. \cite{Malcolm2017MAORISICRA} segment maps by calculating distance image from them, in which each pixel has a pixel value that represents the size of the region it belongs to, i.e. the distance to its closest obstacle.  Then they 
	 merge regions with similar values. This method helps to relieve over-segmentation on corridors. 

	 The main advantage of our method is, that we segment the map based on the topology of the Voronoi graph, which, as discussed in the introduction, thus represents the topology of the environment best. 
	 
	We also use the alpha shape based room segmentation to overcome the oversegmentation problem of our VD-based approach.

	\subsection{Comparison Experiments\label{subsec:experiments_seg}}
	
	This experiment compares our algorithm with Mielle's MAORIS method \cite{Malcolm2017MAORISICRA} and the traditional Voronoi-based segmentation implemented by Bormann \cite{bormann2016room}. 
	We used the best parameter described in their papers for the other two segmentation algorithms. Since their parameters are fixed, to be fair, we used $W= 1.25m$ to do the comparison. We also show the results for the $W$ that are set according to the strategy described in \ref{subsec:experiment_alpha} to show its advantage.
	
	We ran MAORIS, Bormann's Voronoi segmentation and our method on Bormann's sub-dataset, including not only simple maps but also furnitured and incomplete maps.
	Because we get rid of furniture and noise for each input map in the preprocessing step, to be fair, we ran all the methods on the map with the  removal preprocessing. Both of the implementations of our method and the dataset we used are available online\footnote{\url{AvailableInTheFinalVersion}}.
	
\begin{figure}[t]
	\subfigure{\includegraphics[width=0.99\linewidth]{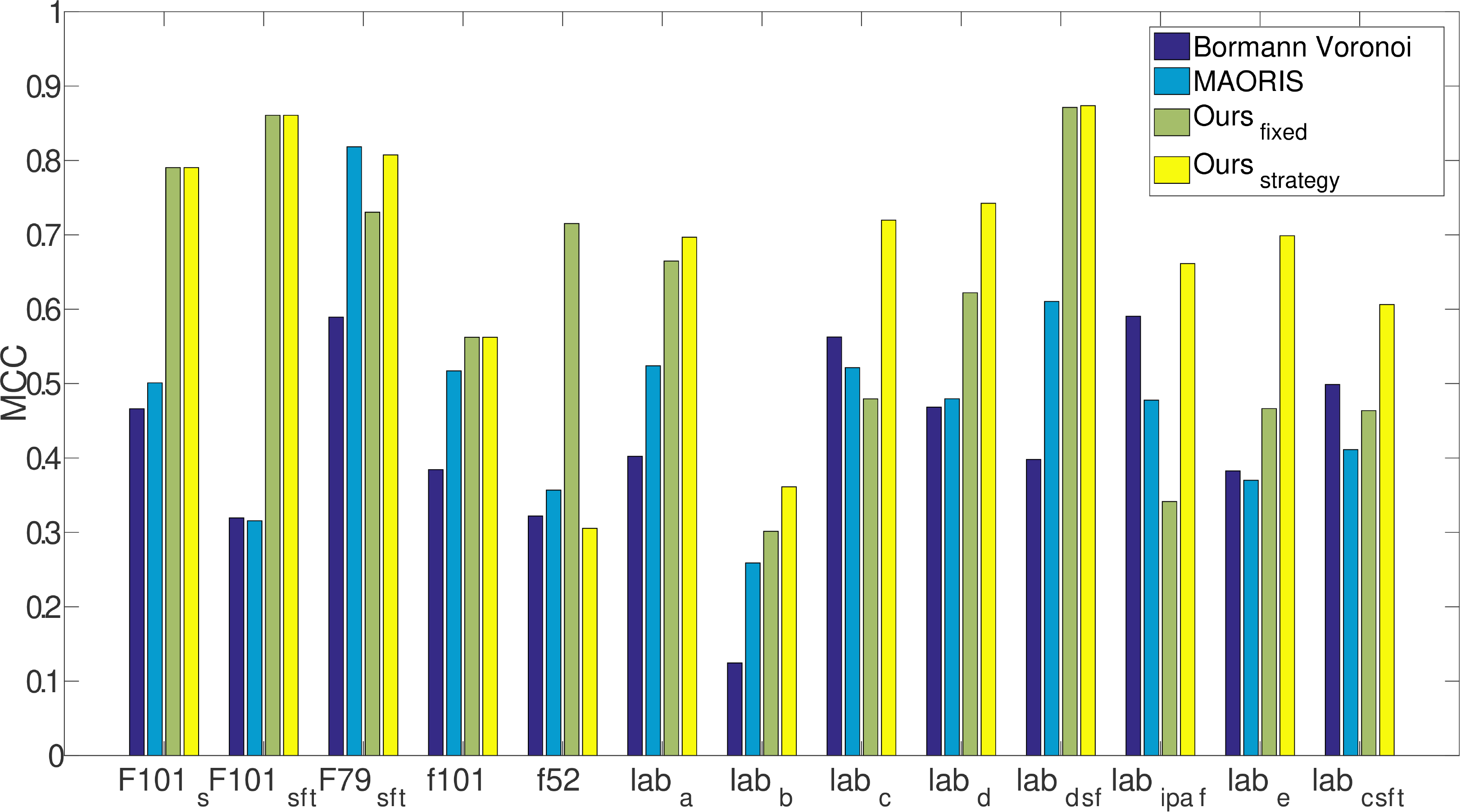}}
	\caption{Compare the segmentation results from three methods.\label{fig:bars}}
\end{figure}

\begin{figure*}[t]
\renewcommand{\thesubfigure}{\space} \makeatletter
\renewcommand{\@thesubfigure}{\space}
\renewcommand{\p@subfigure}{\thefigure} \makeatother
		\subfigure{
		\includegraphics[width=0.15\linewidth]{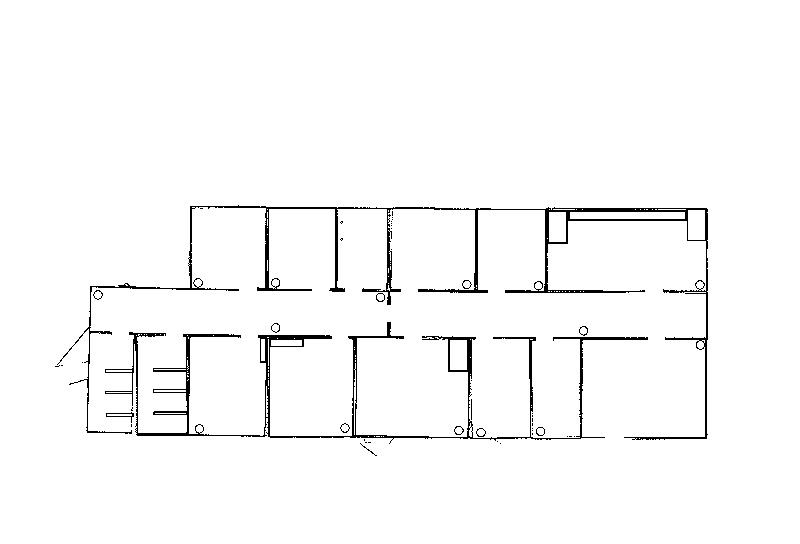}}
		\subfigure{ \includegraphics[width=0.15\linewidth]{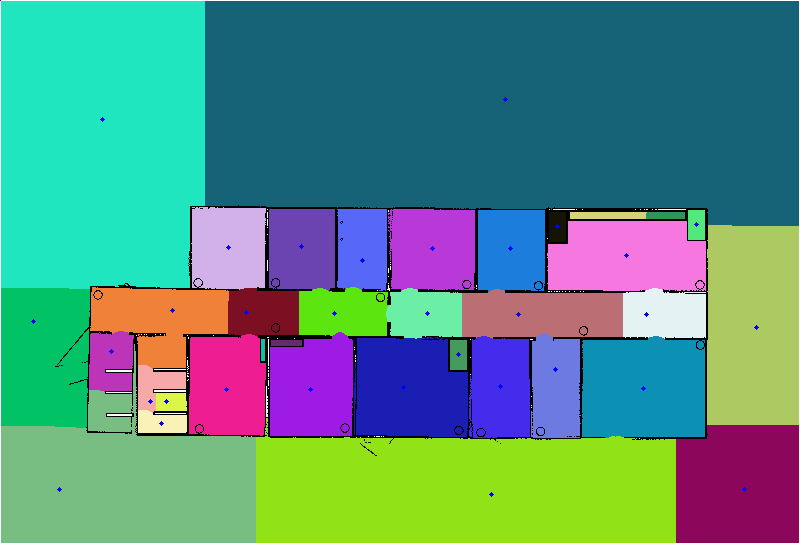}}
		\subfigure{ \includegraphics[width=0.15\linewidth]{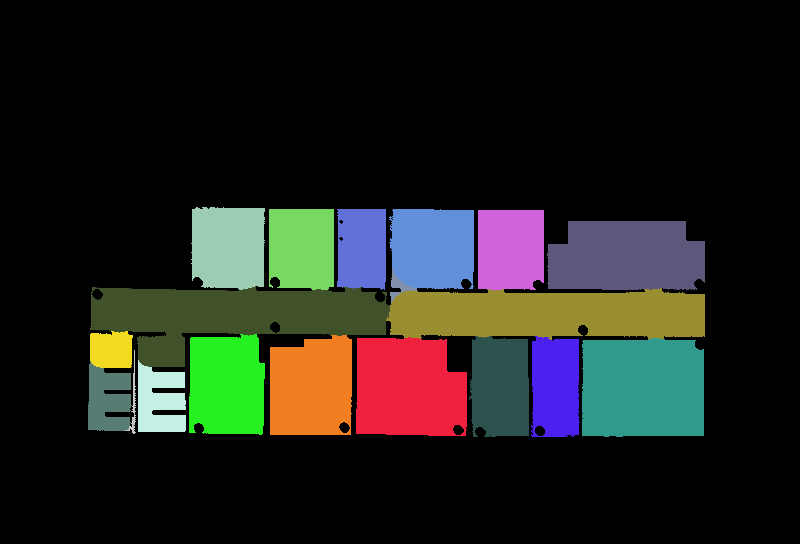}}
		\subfigure{ \includegraphics[width=0.15\linewidth]{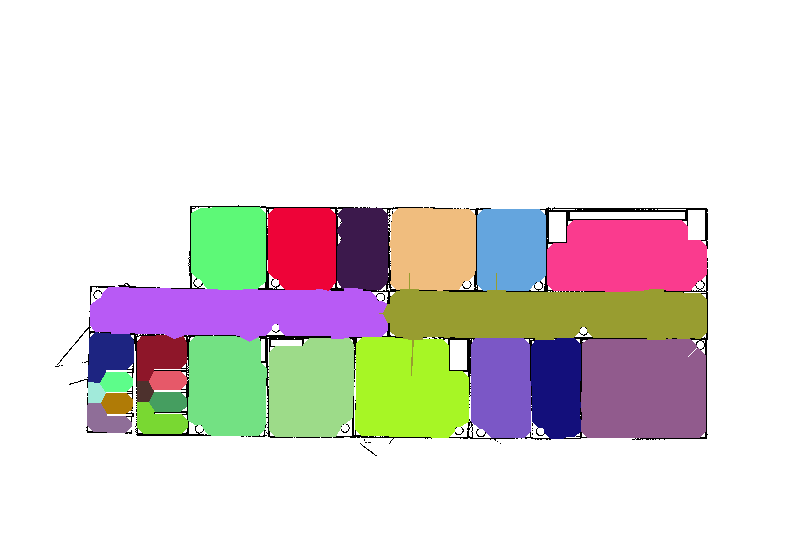}}
		\subfigure{ \includegraphics[width=0.15\linewidth]{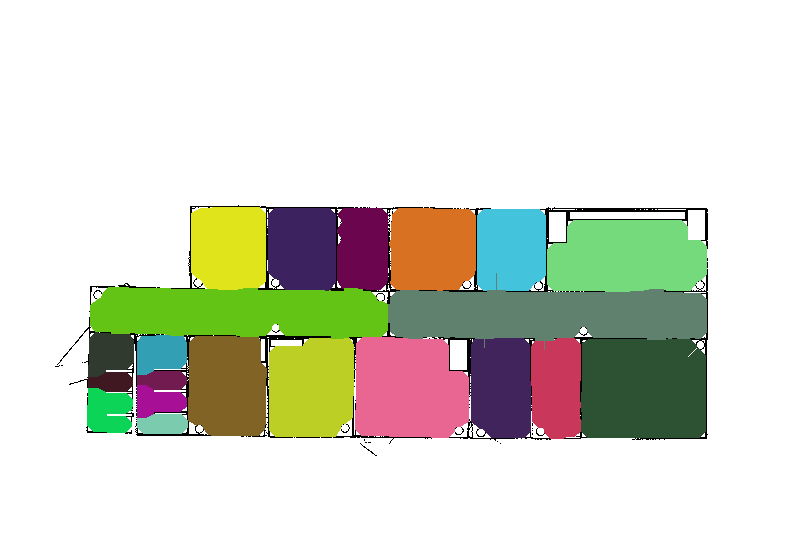}}
		\subfigure{ \includegraphics[width=0.15\linewidth]{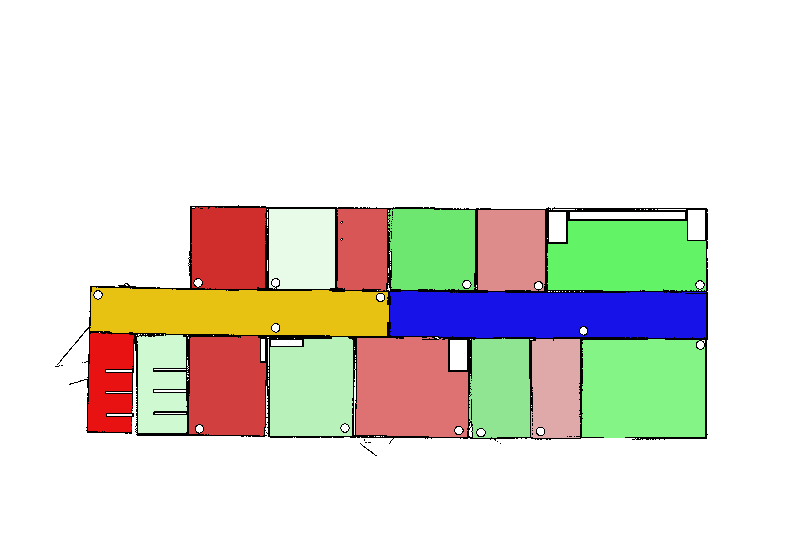}}
			
		\subfigure{				\includegraphics[width=0.15\linewidth]{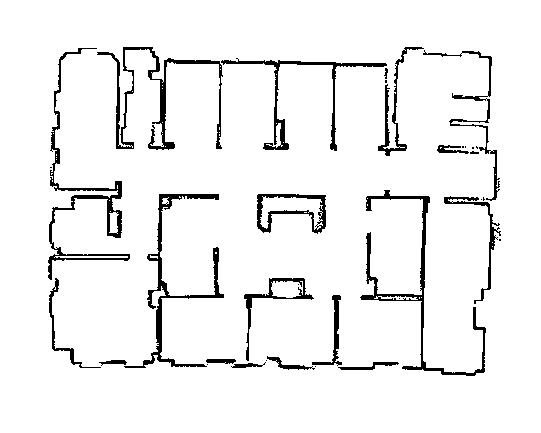}}
		\subfigure{ \includegraphics[width=0.15\linewidth]{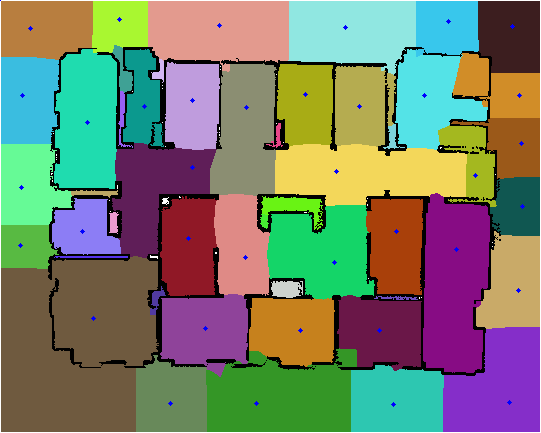}}
		\subfigure{ \includegraphics[width=0.15\linewidth]{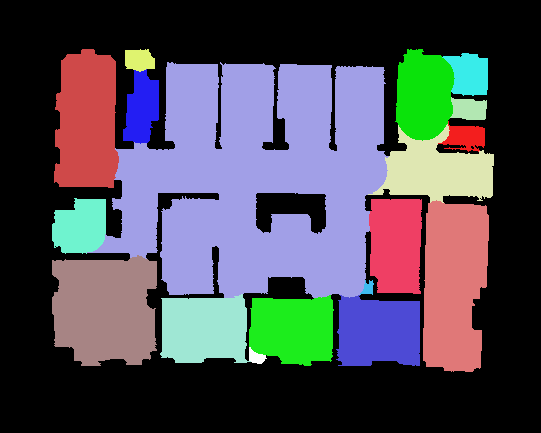}}
		\subfigure{ \includegraphics[width=0.15\linewidth]{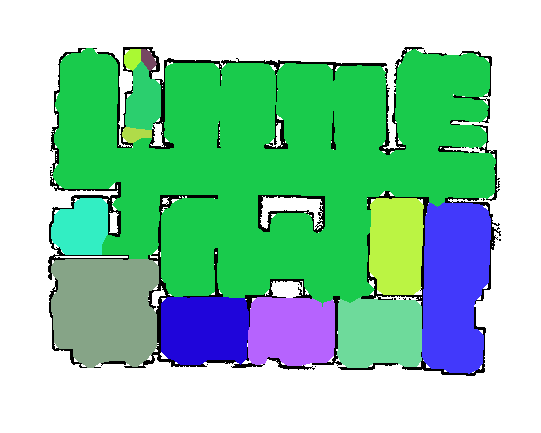}}
		\subfigure{ \includegraphics[width=0.15\linewidth]{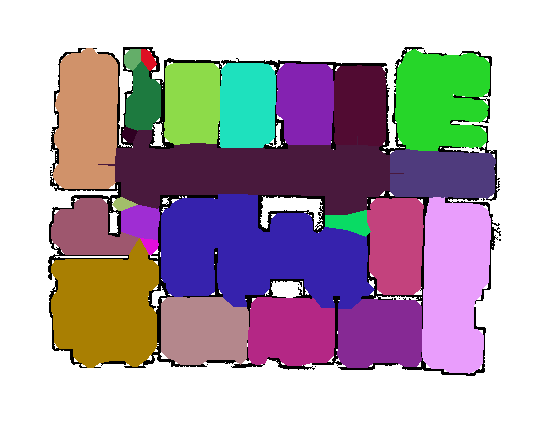}}
		\subfigure{ \includegraphics[width=0.15\linewidth]{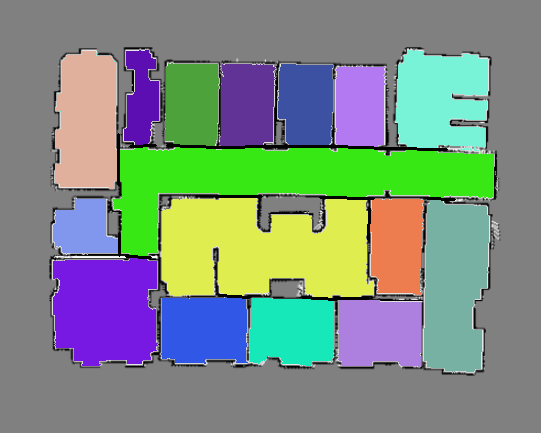}}
		
		\subfigure{				\includegraphics[width=0.15\linewidth]{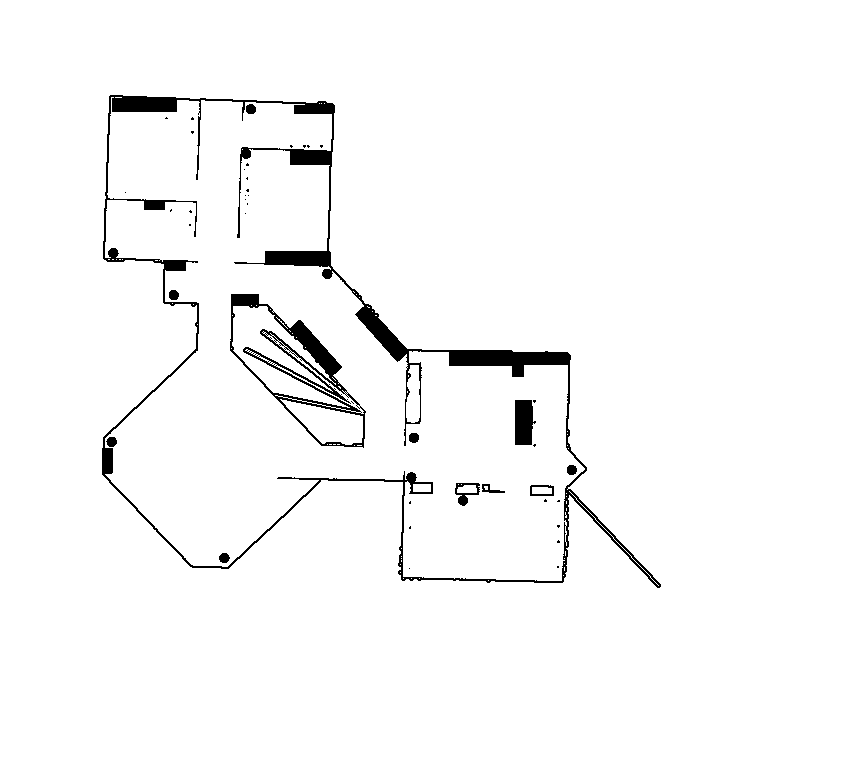}}
		\subfigure{ \includegraphics[width=0.15\linewidth]{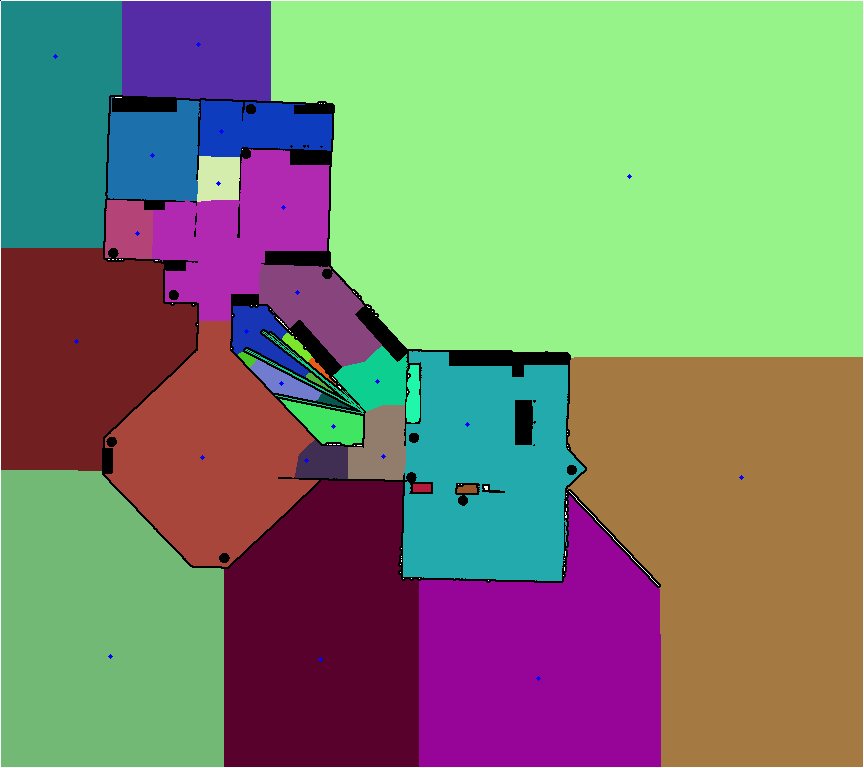}}
		\subfigure{ \includegraphics[width=0.15\linewidth]{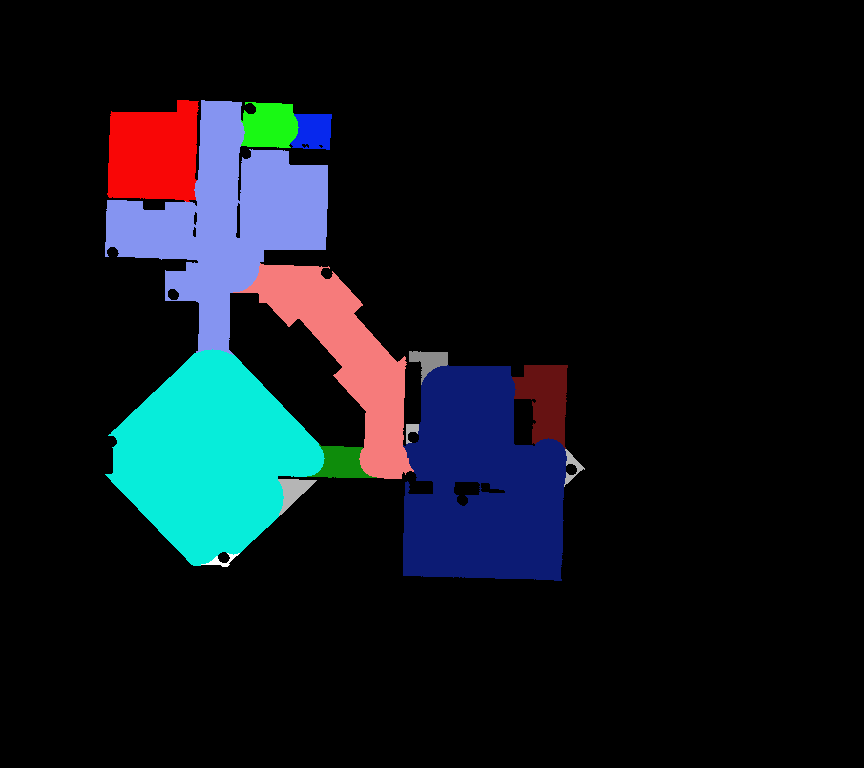}}
		\subfigure{ \includegraphics[width=0.15\linewidth]{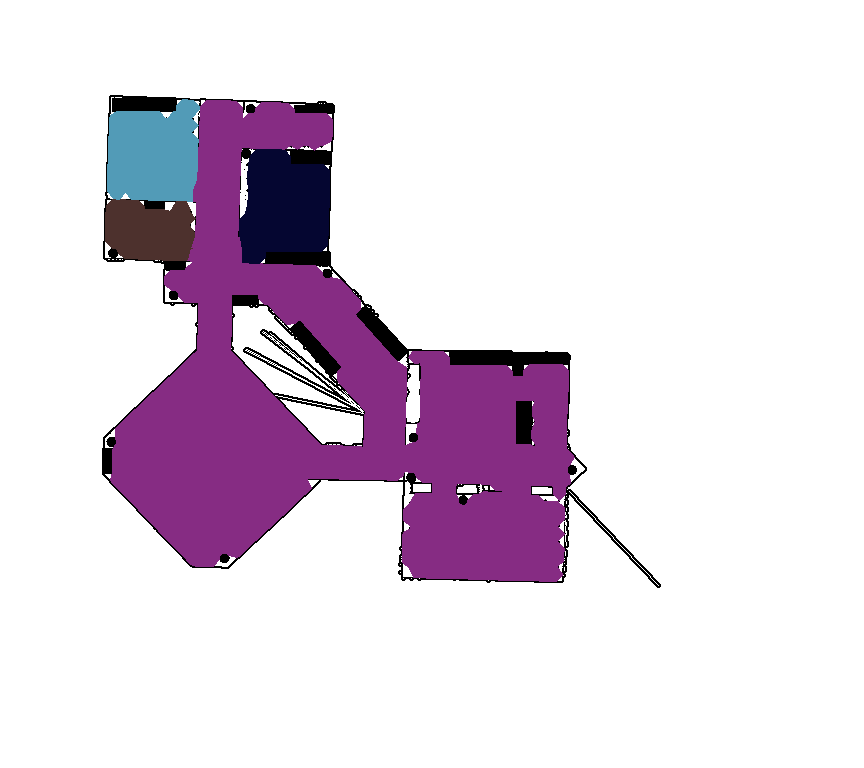}}
		\subfigure{ \includegraphics[width=0.15\linewidth]{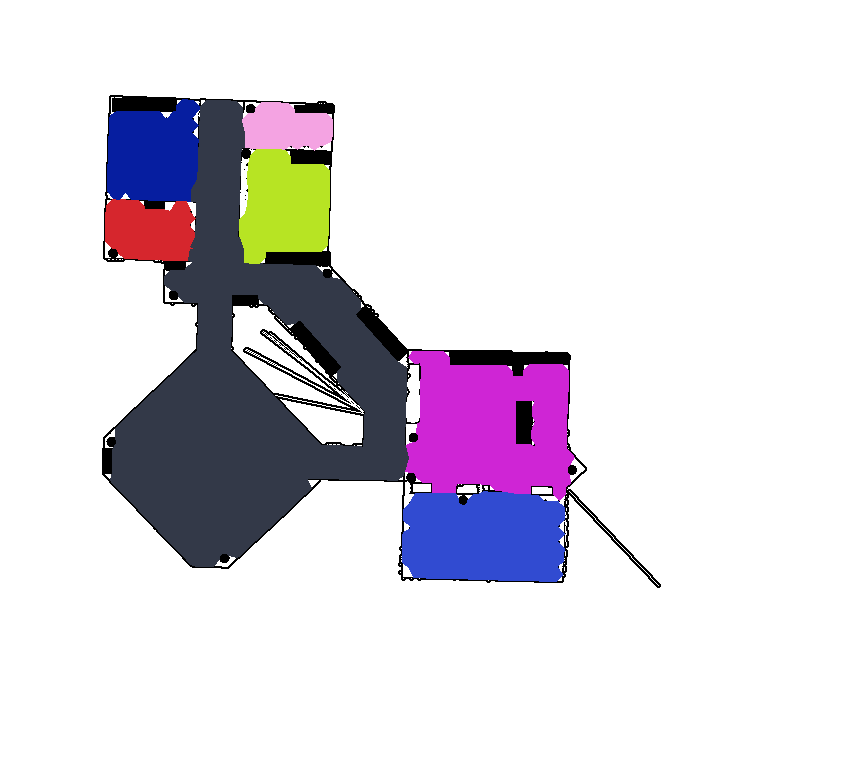}}
		\subfigure{ \includegraphics[width=0.15\linewidth]{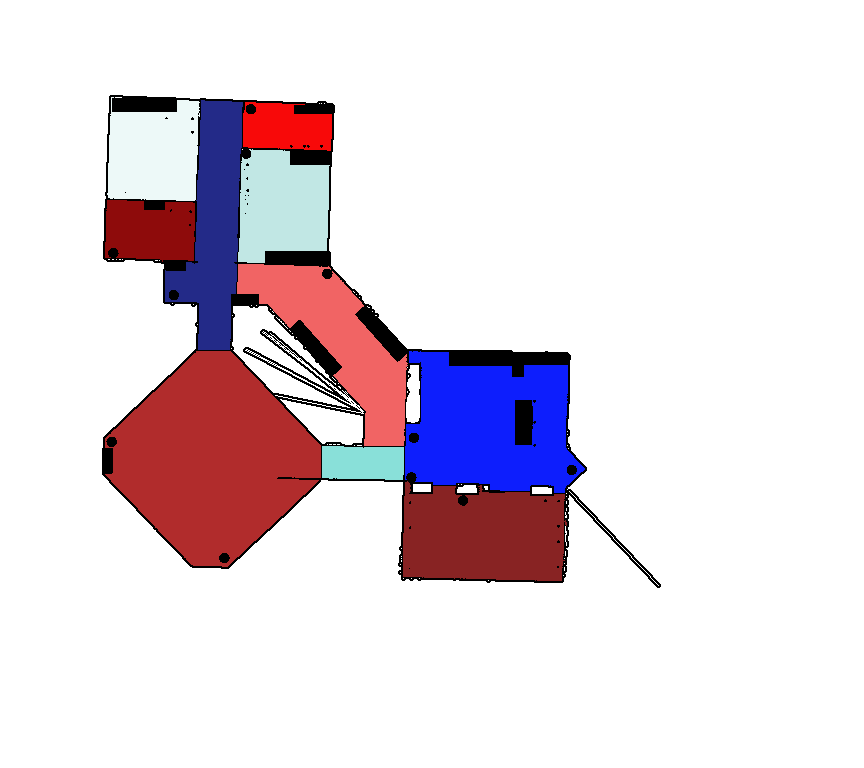}}
		
			\subfigure{				\includegraphics[width=0.15\linewidth]{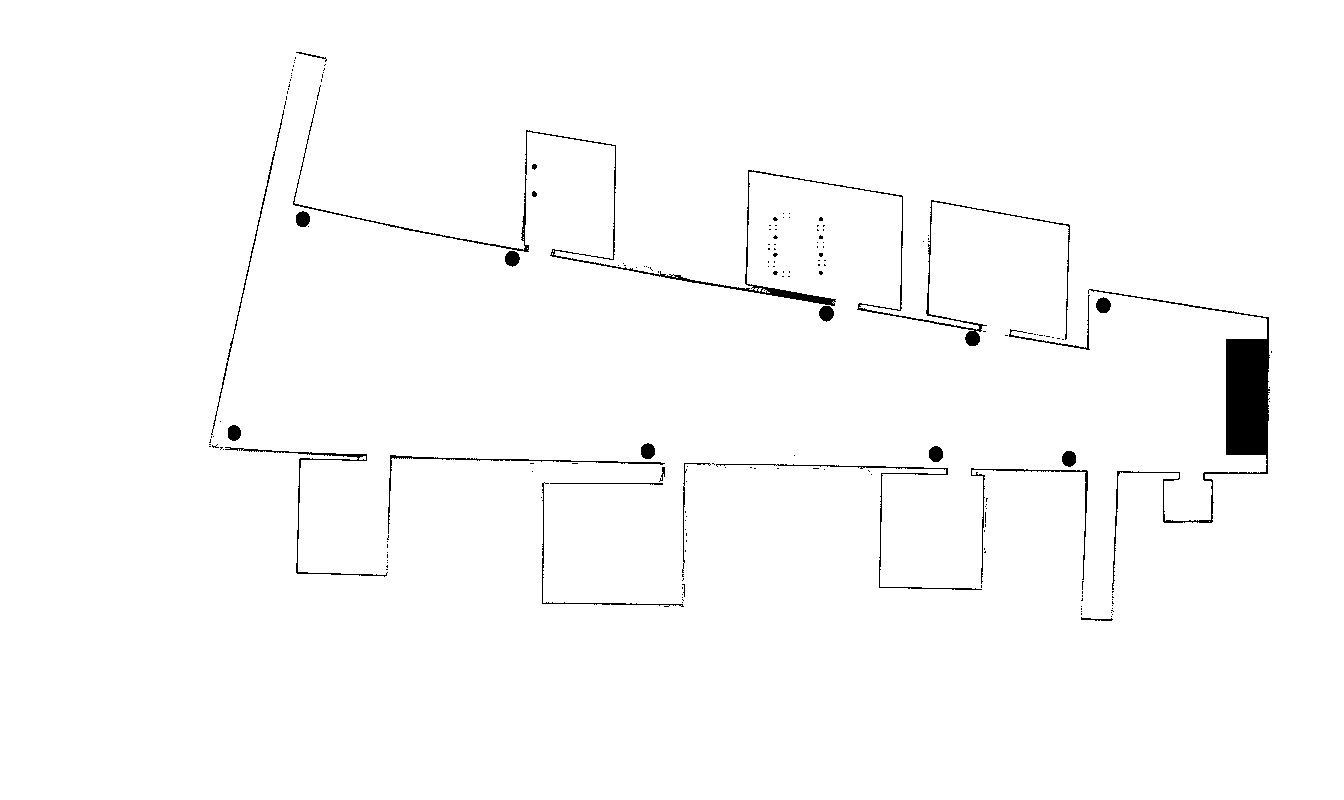}}
			\subfigure{
	    	\includegraphics[width=0.15\linewidth]{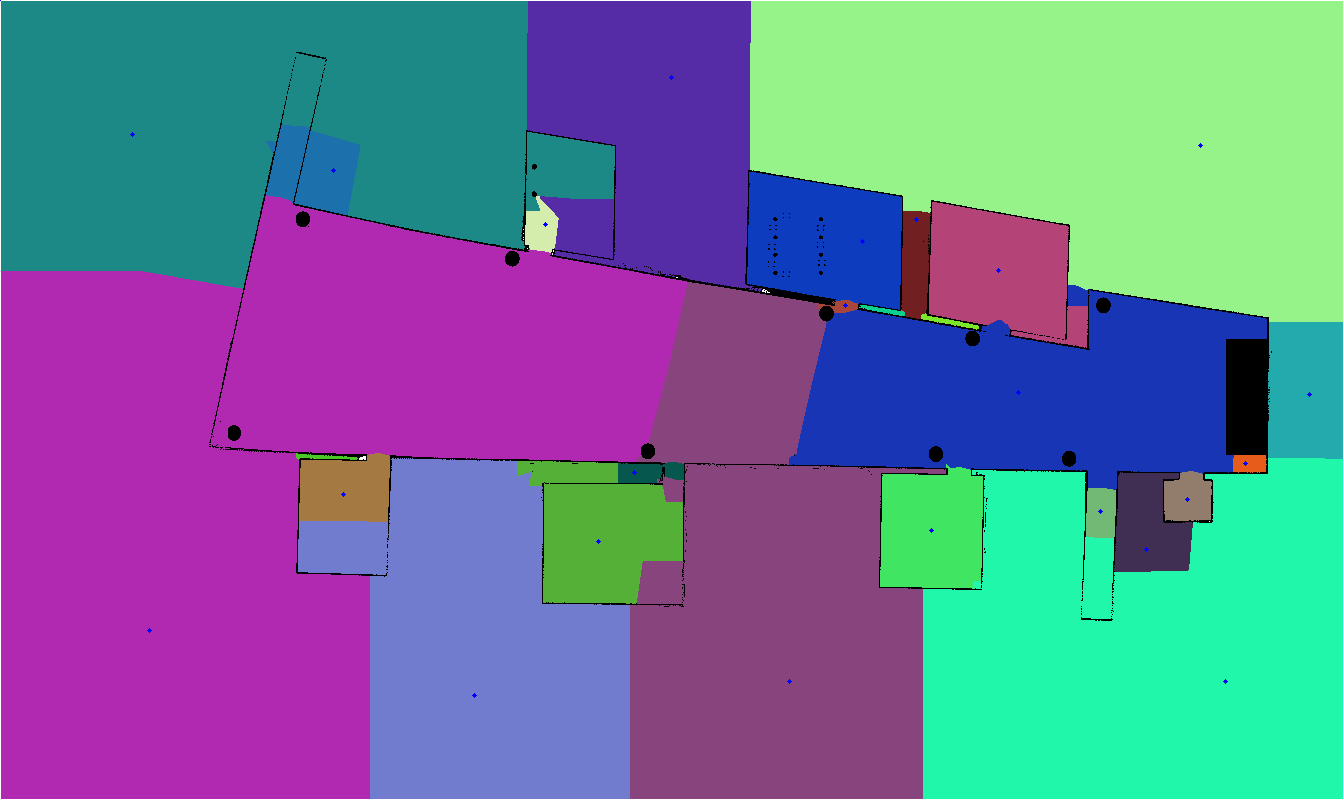}}
			\subfigure{	\includegraphics[width=0.15\linewidth]{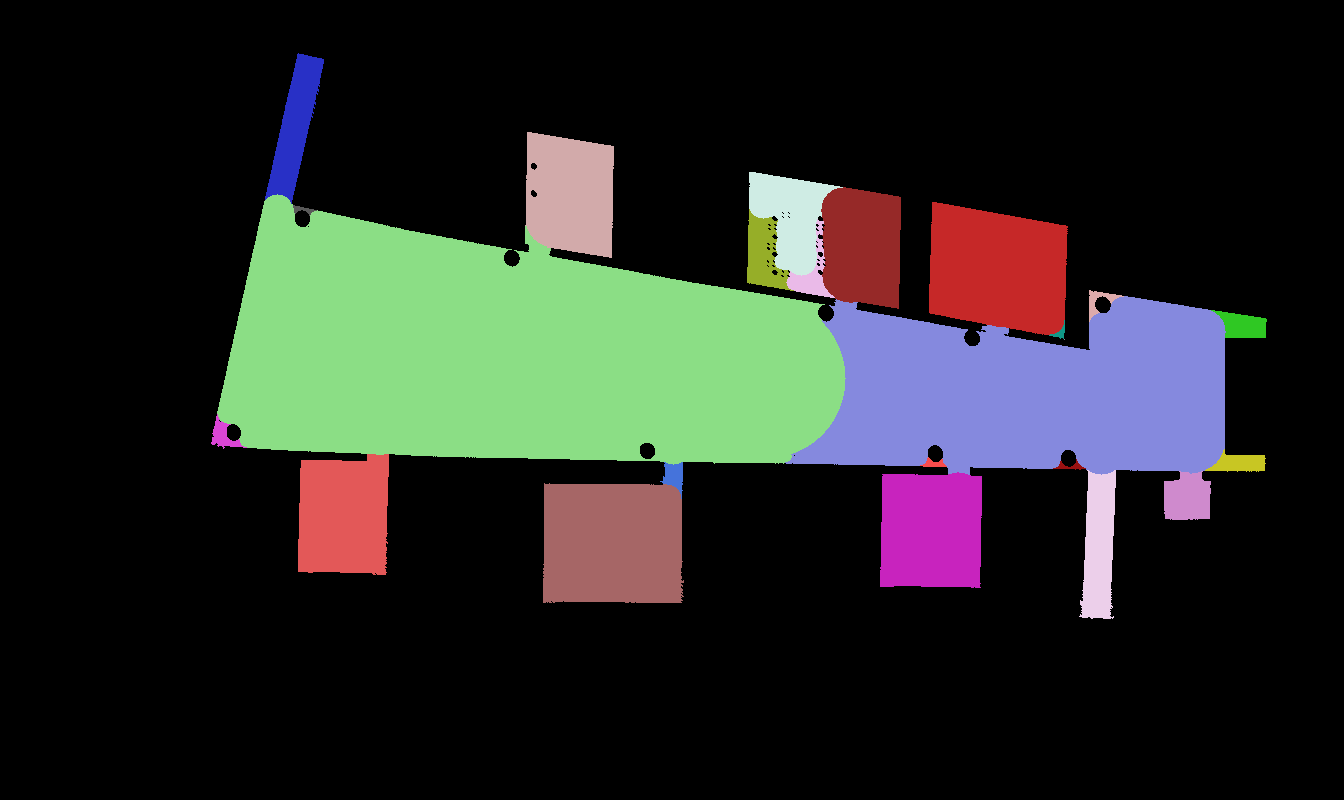}}
			\subfigure{
		   	\includegraphics[width=0.15\linewidth]{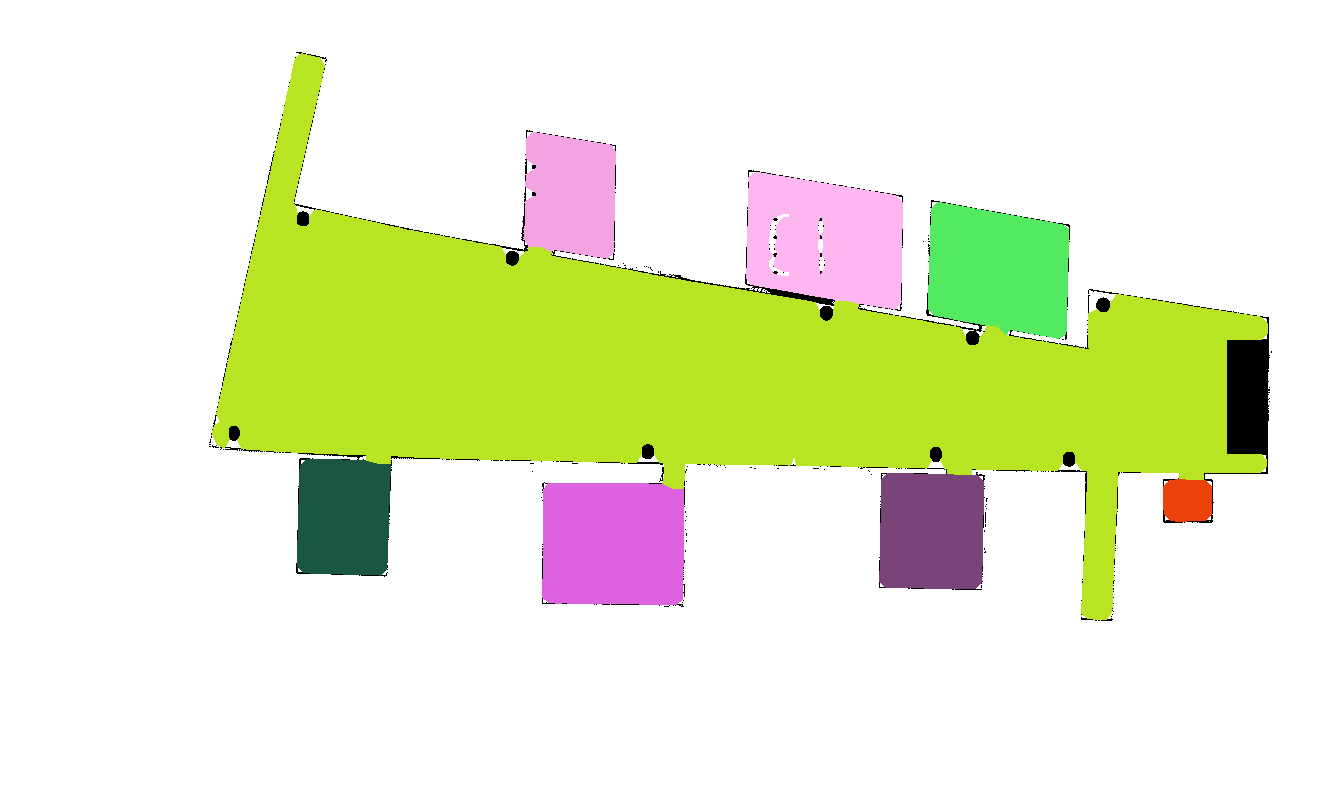}}
			\subfigure{
		   	\includegraphics[width=0.15\linewidth]{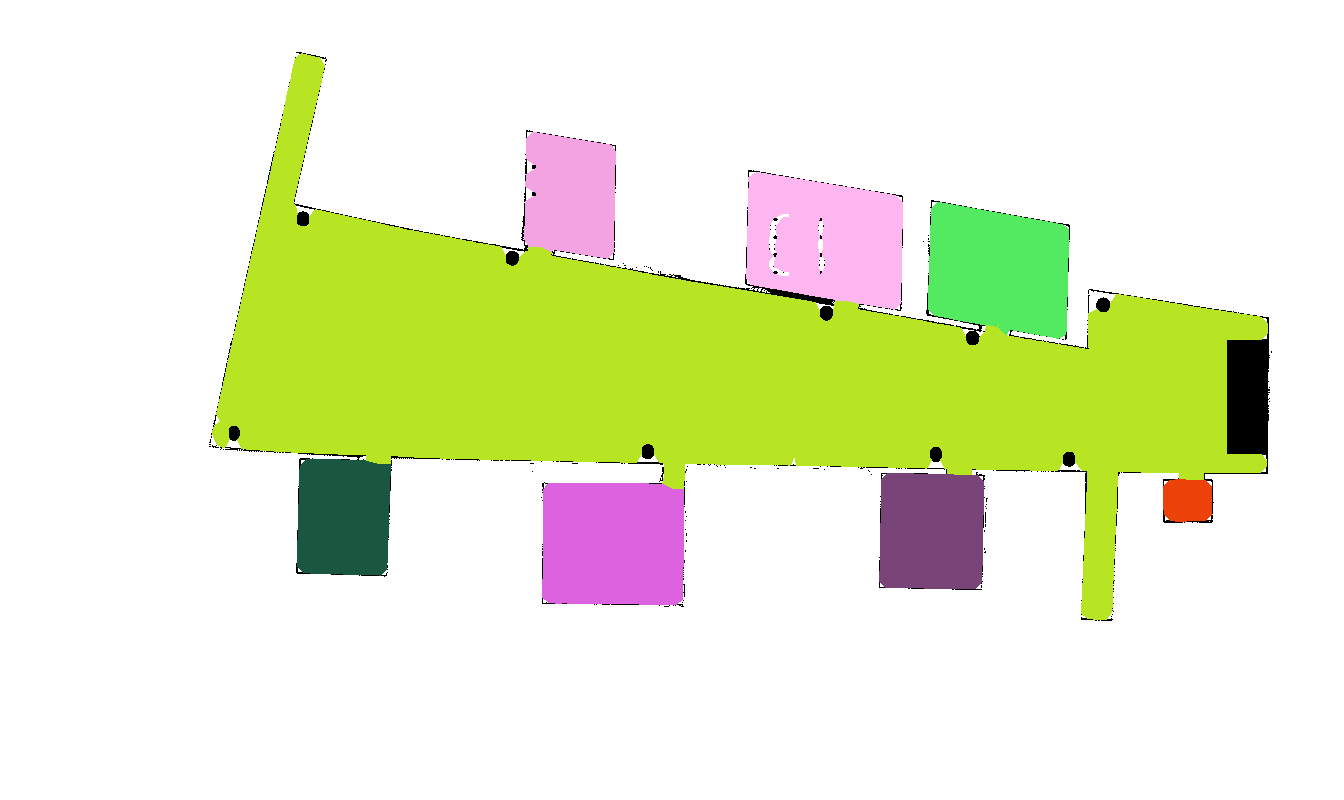}}
			\subfigure{			\includegraphics[width=0.15\linewidth]{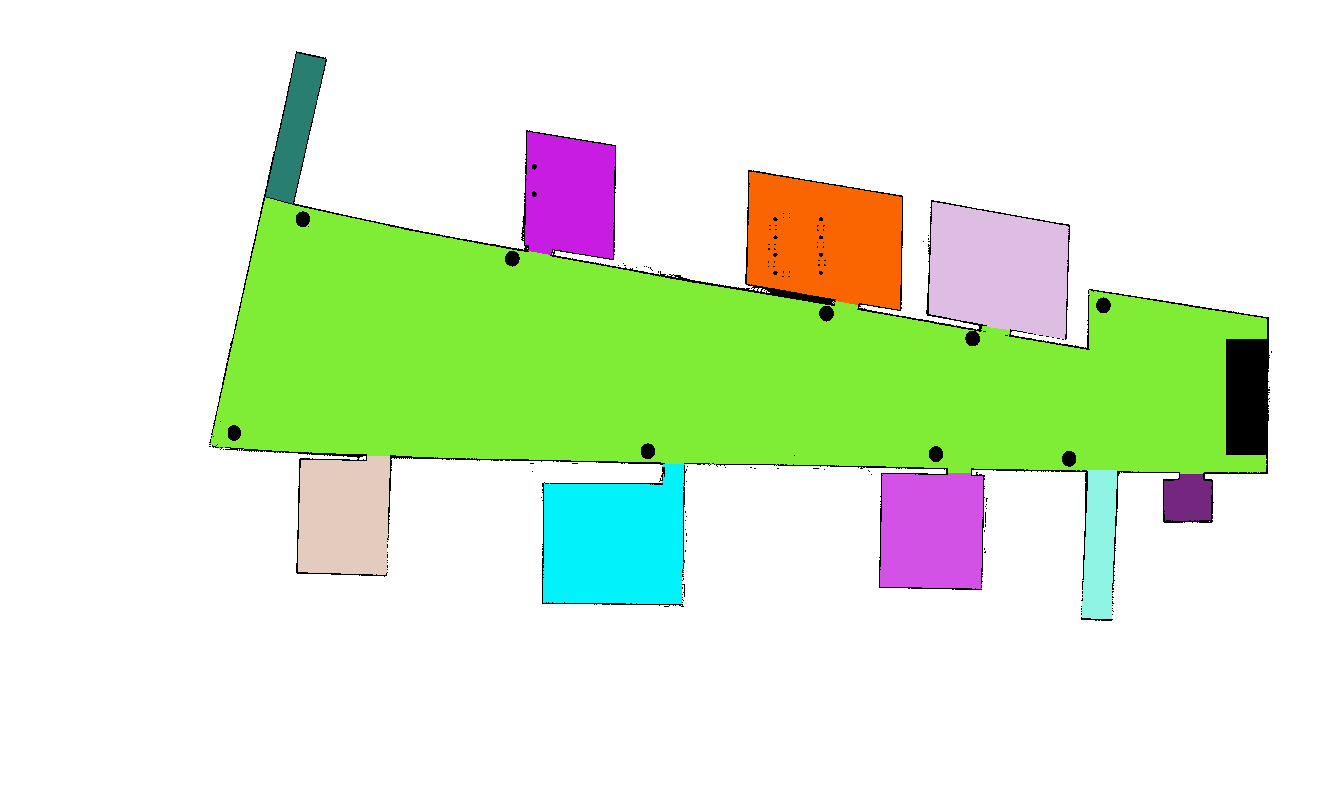}}
			
		\subfigure{	\includegraphics[width=0.15\linewidth]{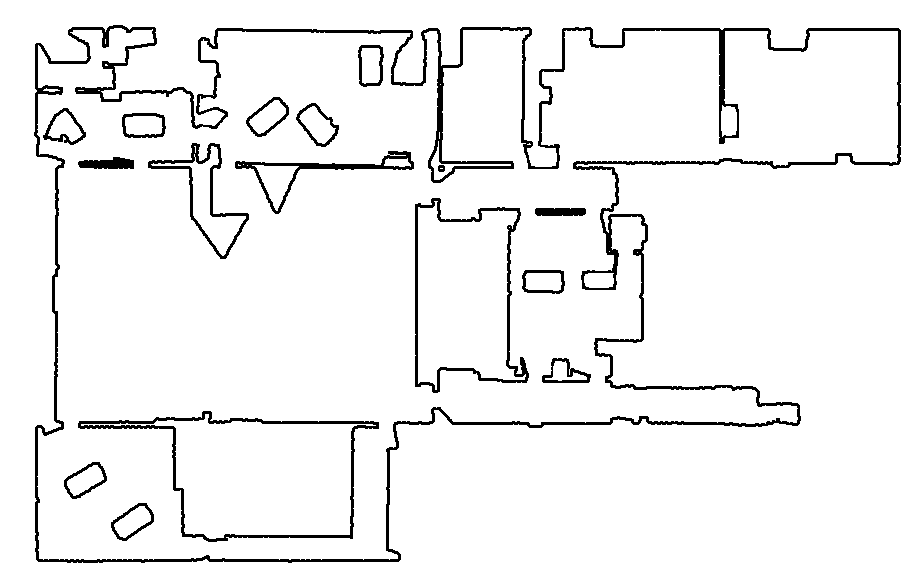}}
		\subfigure{ \includegraphics[width=0.15\linewidth]{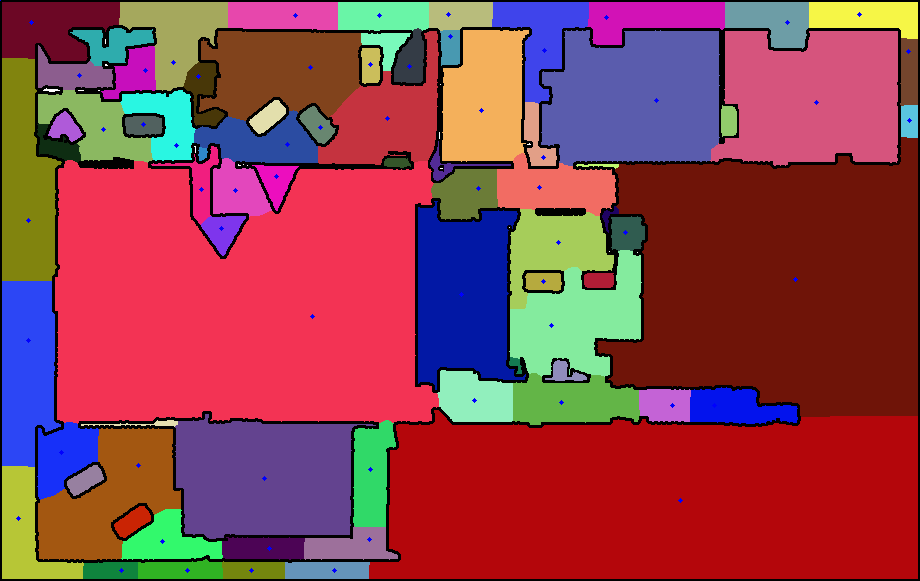}}
		\subfigure{	\includegraphics[width=0.15\linewidth]{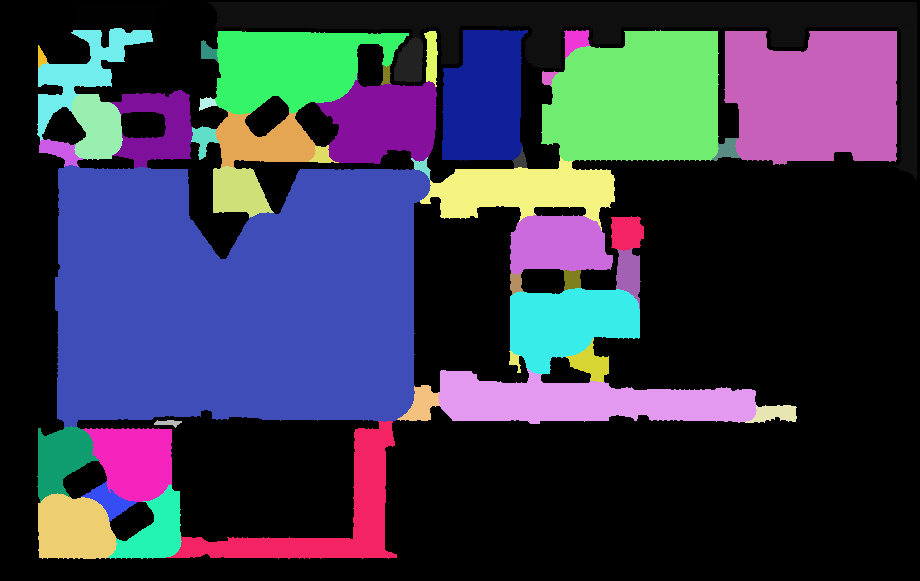}}
		\subfigure{ \includegraphics[width=0.15\linewidth]{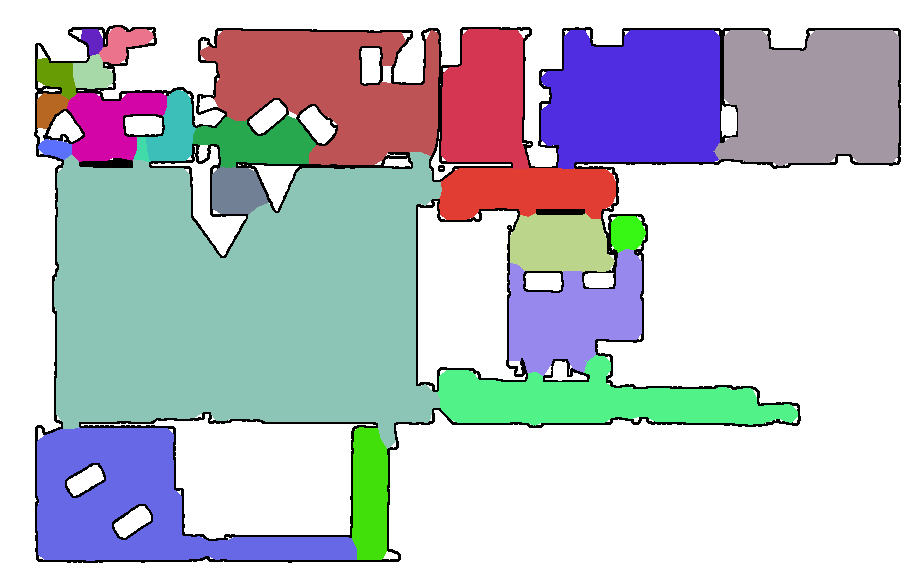}}
		\subfigure{ \includegraphics[width=0.15\linewidth]{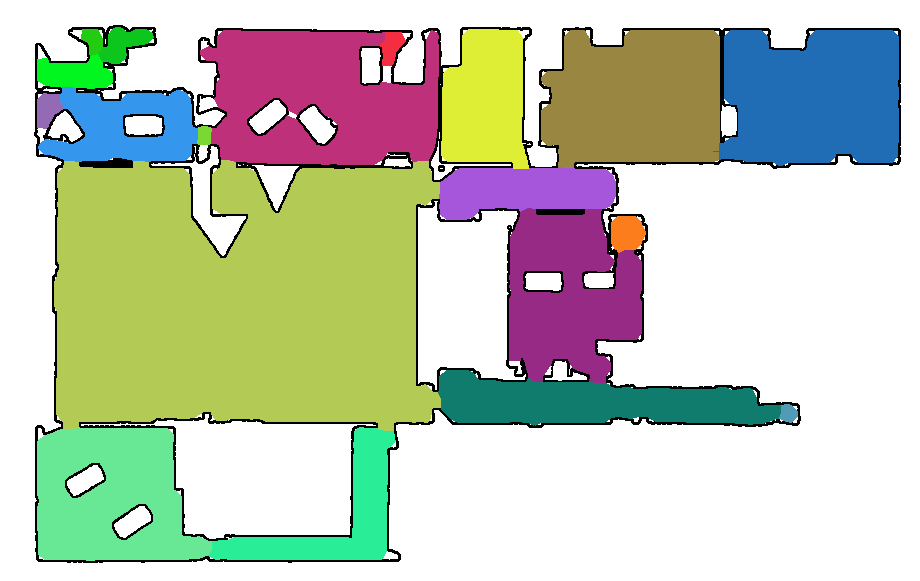}}
		\subfigure{	\includegraphics[width=0.15\linewidth]{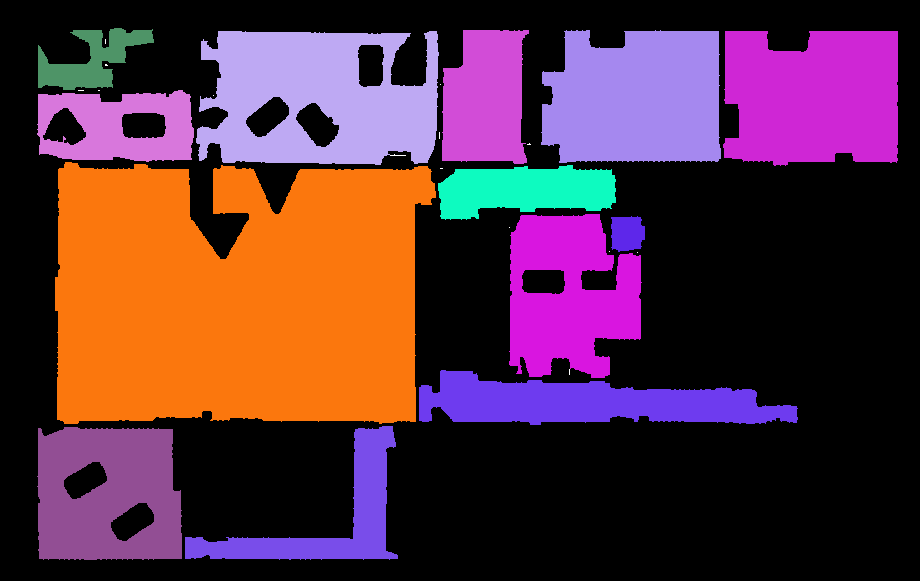}}
		


		\subfigure{	\includegraphics[width=0.15\linewidth]{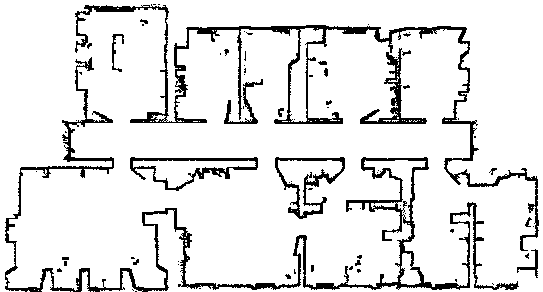}}
		\subfigure{ \includegraphics[width=0.15\linewidth]{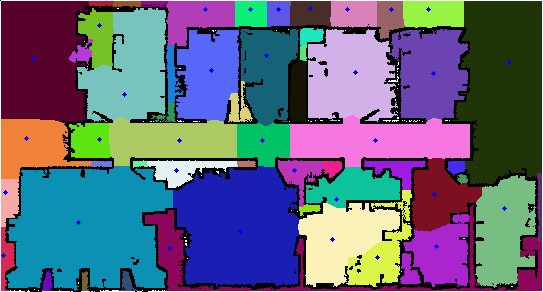}}
		\subfigure{ \includegraphics[width=0.15\linewidth]{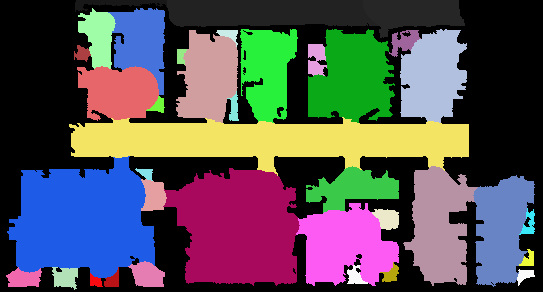}}
		\subfigure{ \includegraphics[width=0.15\linewidth]{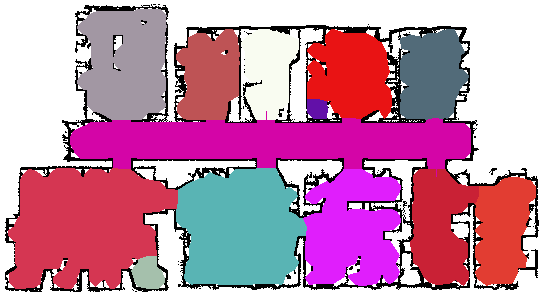}}
		\subfigure{ \includegraphics[width=0.15\linewidth]{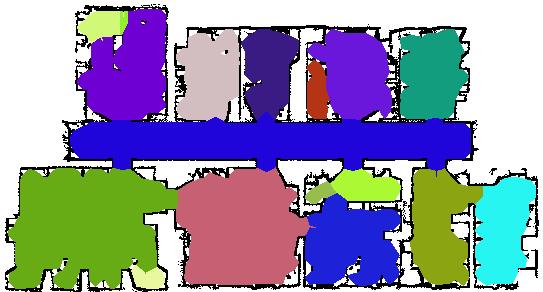}}
		\subfigure{	\includegraphics[width=0.15\linewidth]{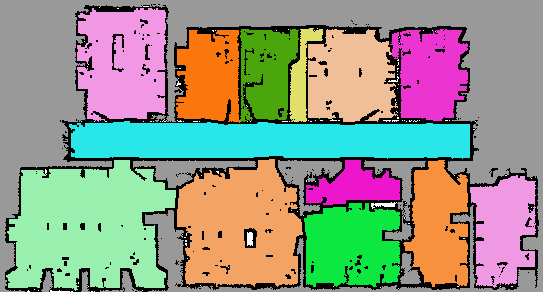}}
		
		\subfigure{	\includegraphics[width=0.15\linewidth]{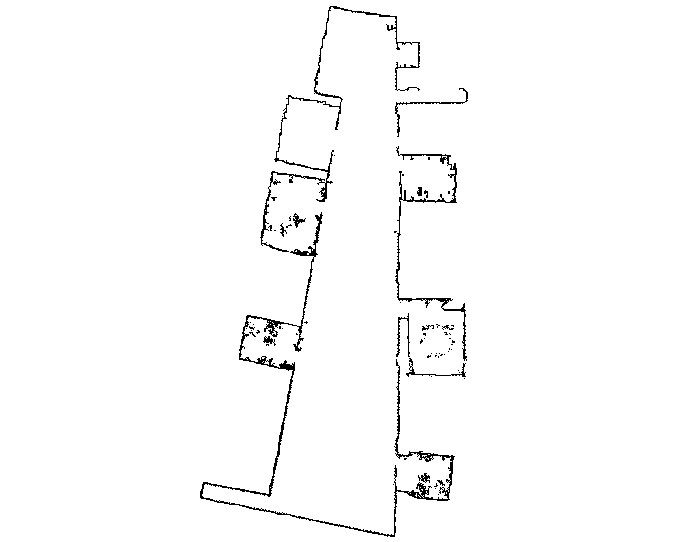}}
		\subfigure{ \includegraphics[width=0.15\linewidth]{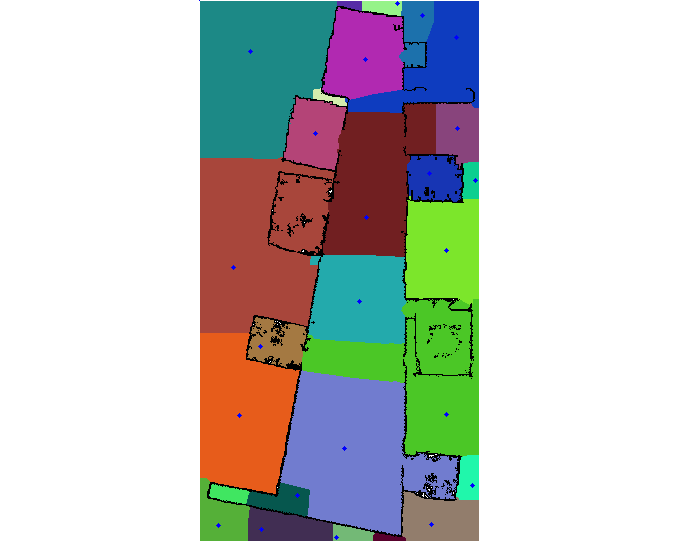}}
		\subfigure{	\includegraphics[width=0.15\linewidth]{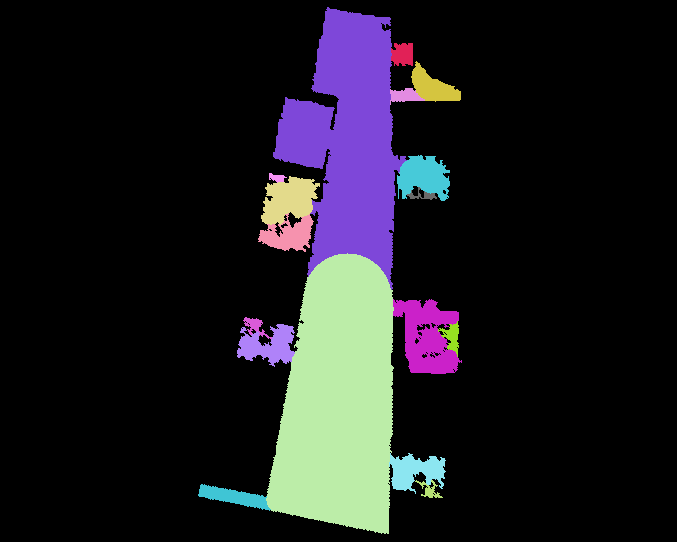}}
		\subfigure{ \includegraphics[width=0.15\linewidth]{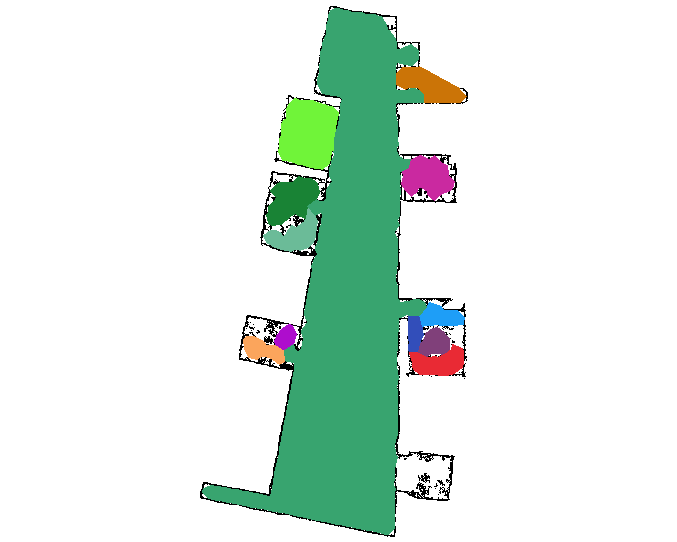}}
		\subfigure{ \includegraphics[width=0.15\linewidth]{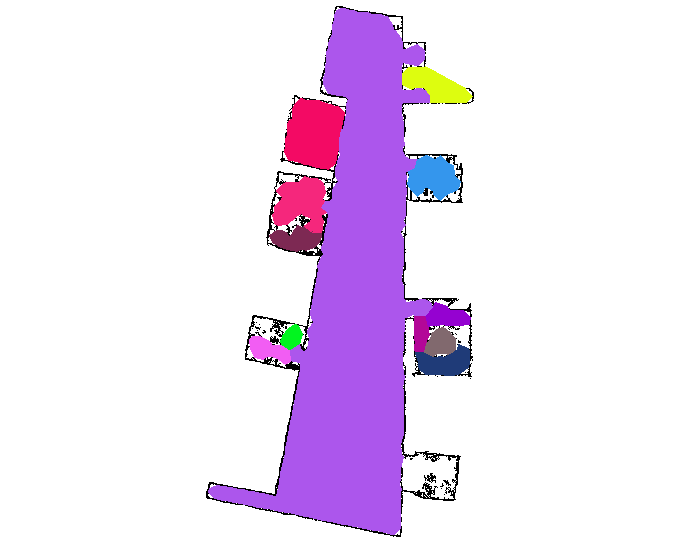}}
		\subfigure{	\includegraphics[width=0.15\linewidth]{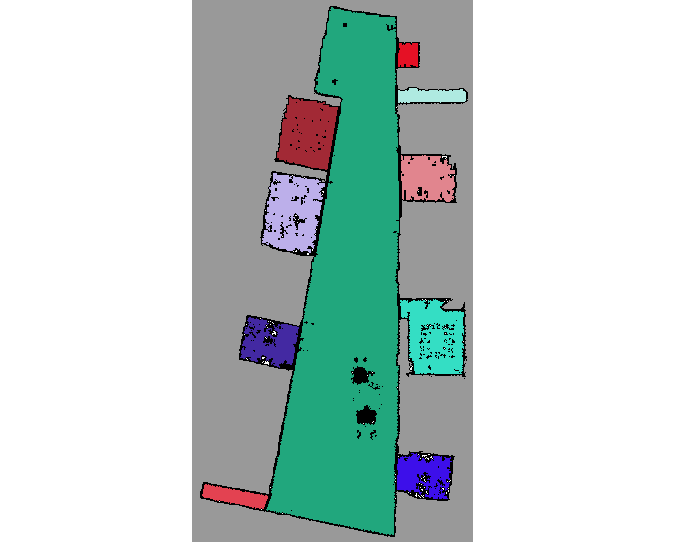}}
		
		\subfigure{ \includegraphics[width=0.15\linewidth]{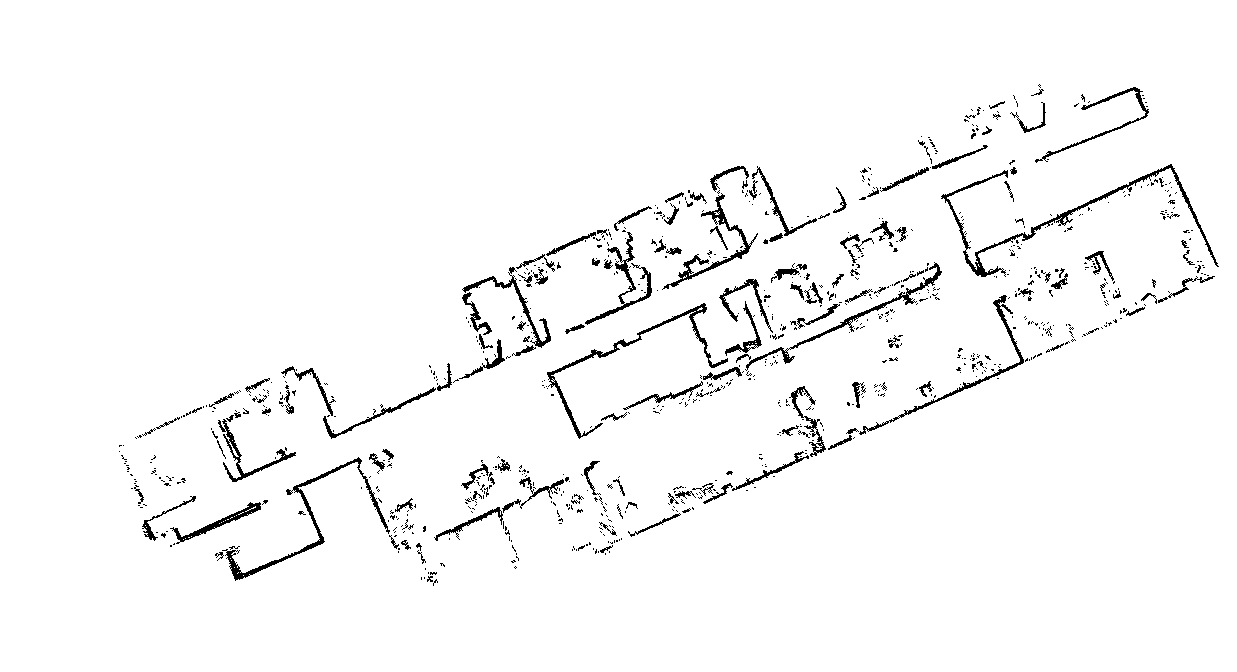}}
		\subfigure{ \includegraphics[width=0.15\linewidth]{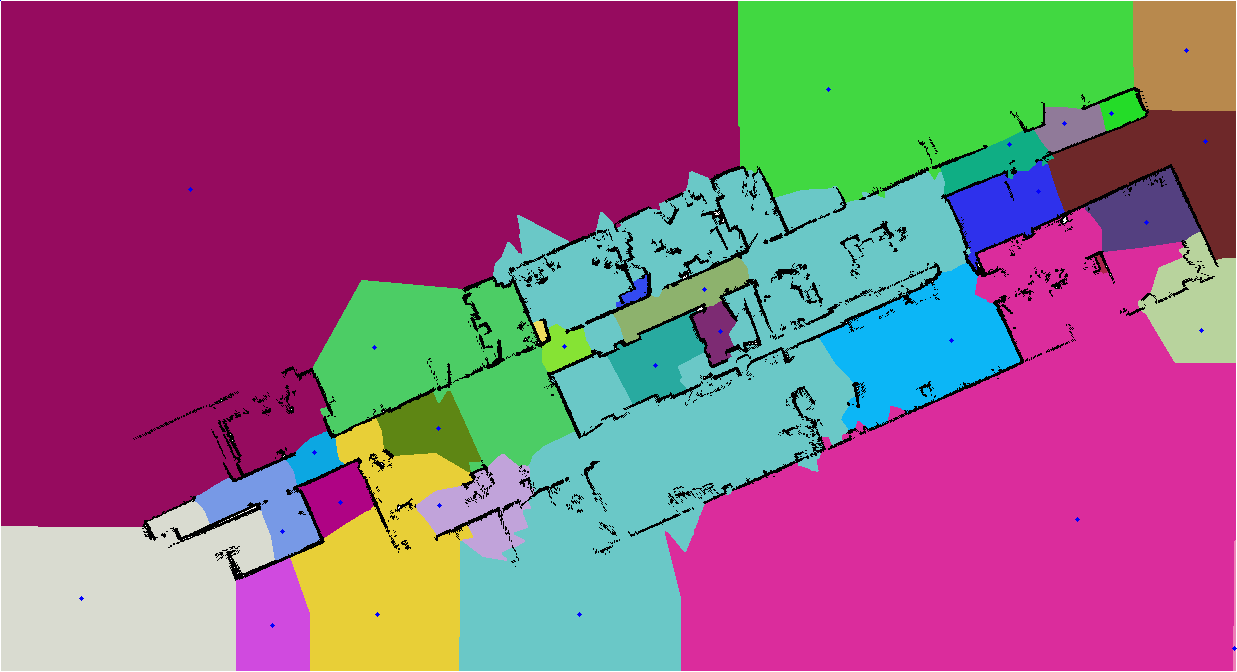}}
		\subfigure{ \includegraphics[width=0.15\linewidth]{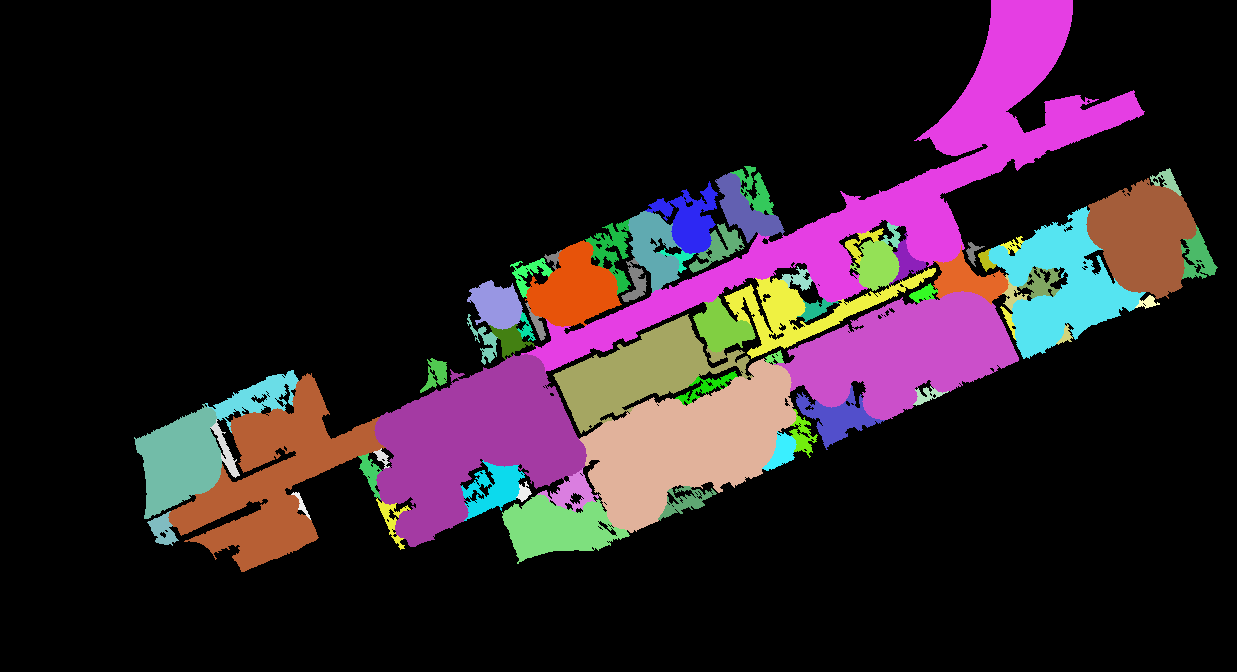}}
		\subfigure{ \includegraphics[width=0.15\linewidth]{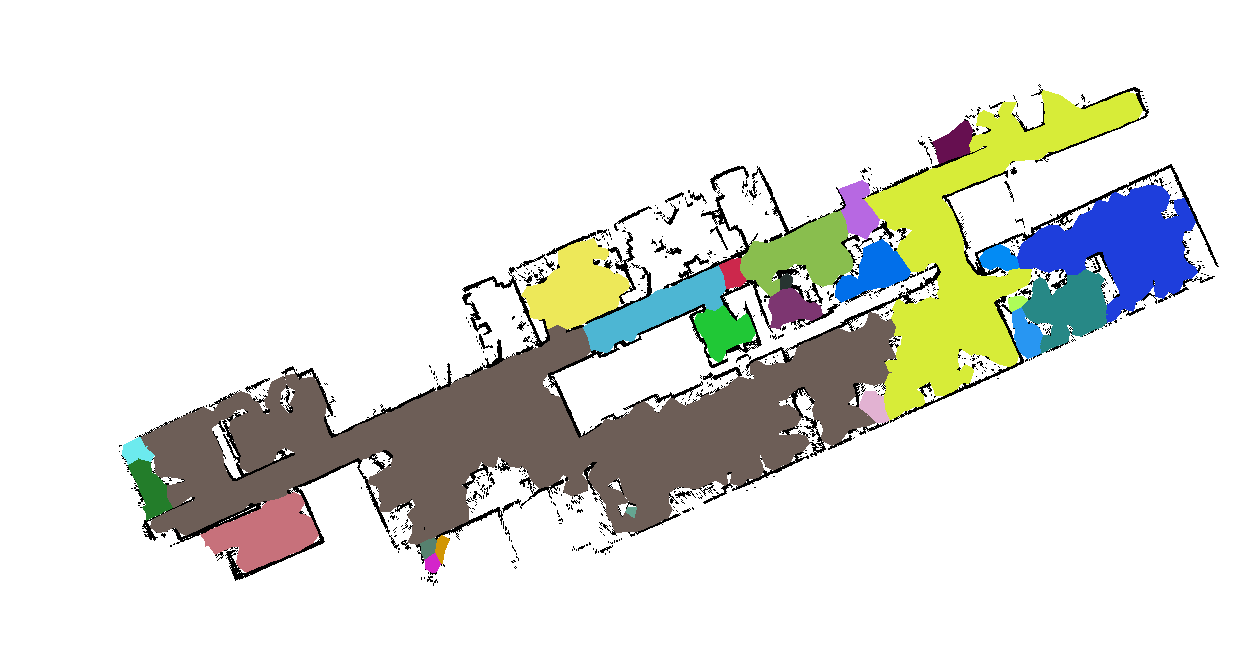}}
		\subfigure{ \includegraphics[width=0.15\linewidth]{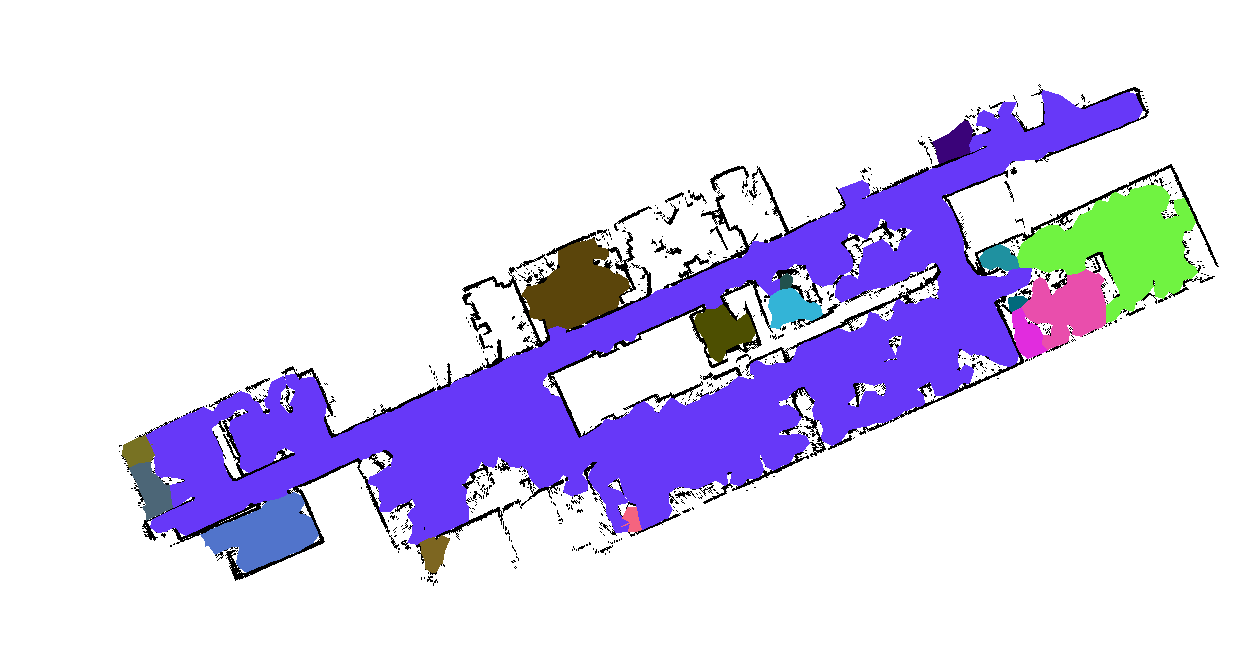}}
		\subfigure{ \includegraphics[width=0.15\linewidth]{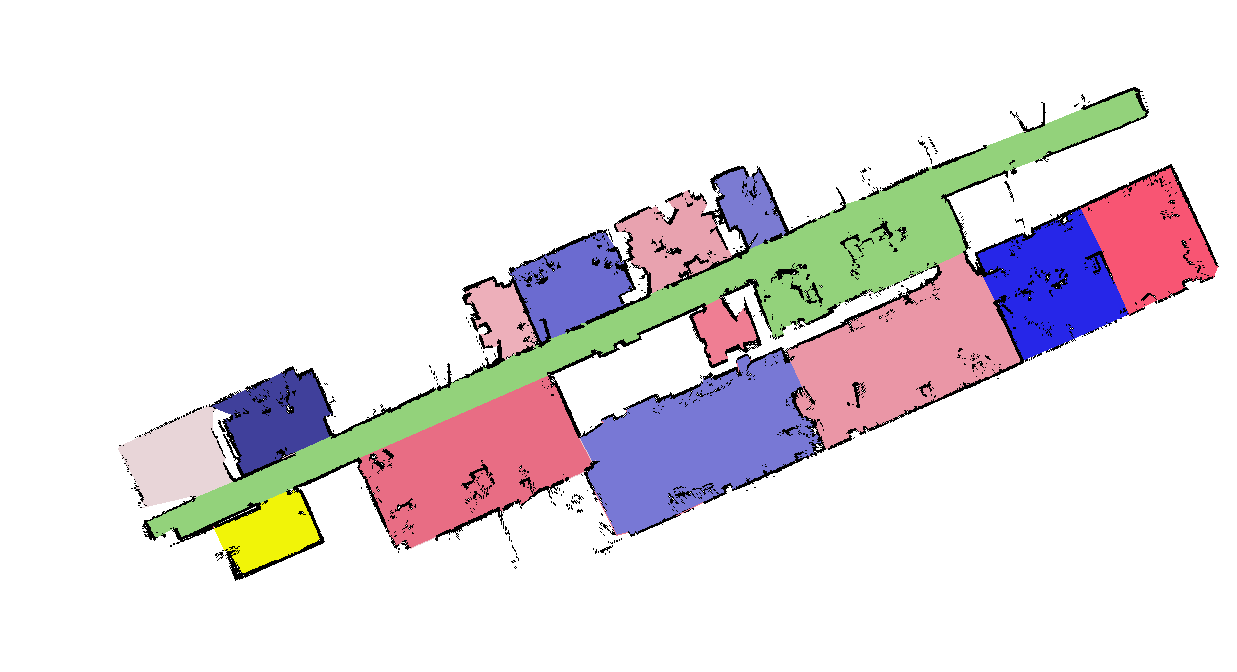}}
		\subfigure[Input]{ \includegraphics[width=0.15\linewidth]{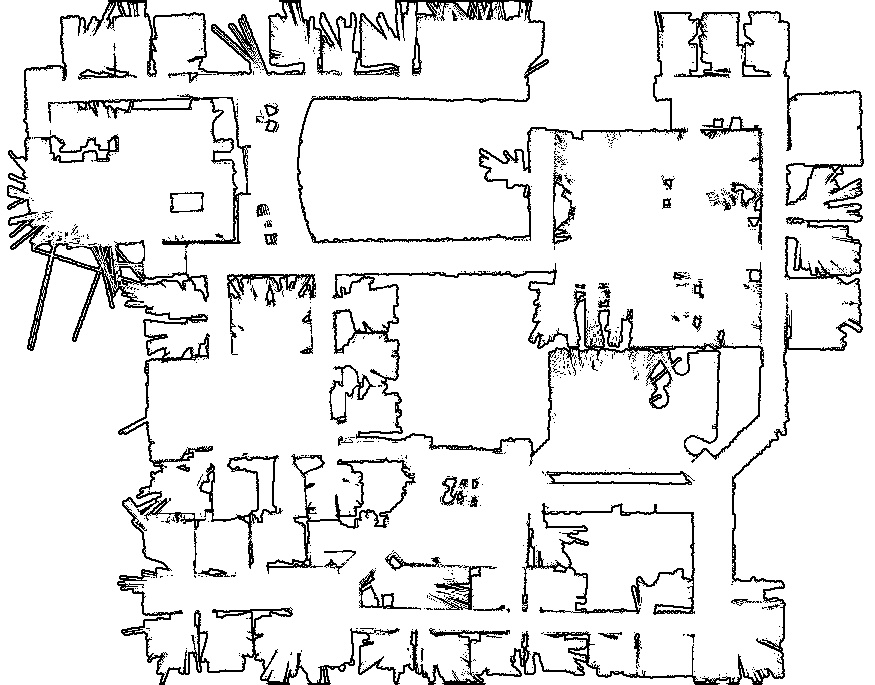}}
		\subfigure[Bormann's]{ \includegraphics[width=0.15\linewidth]{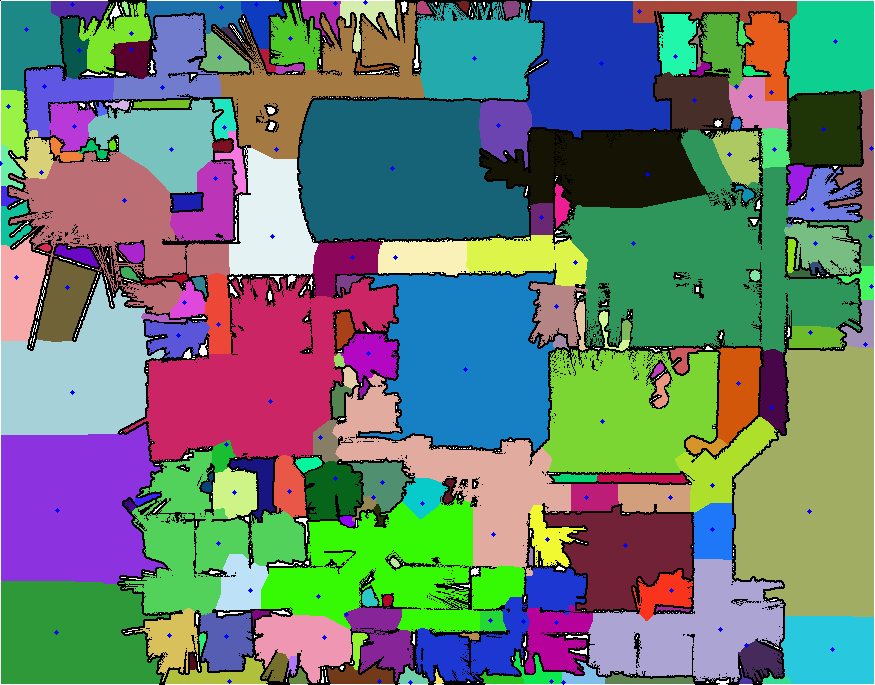}}
		\subfigure[MAORIS]{	\includegraphics[width=0.15\linewidth]{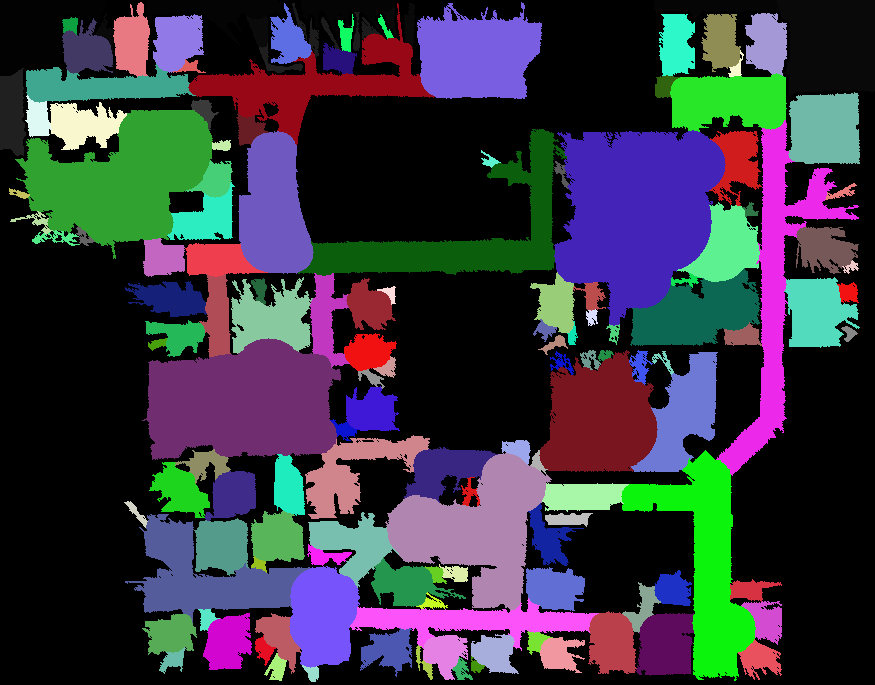}}
		\subfigure[Ours(fixed)]{ \includegraphics[width=0.15\linewidth]{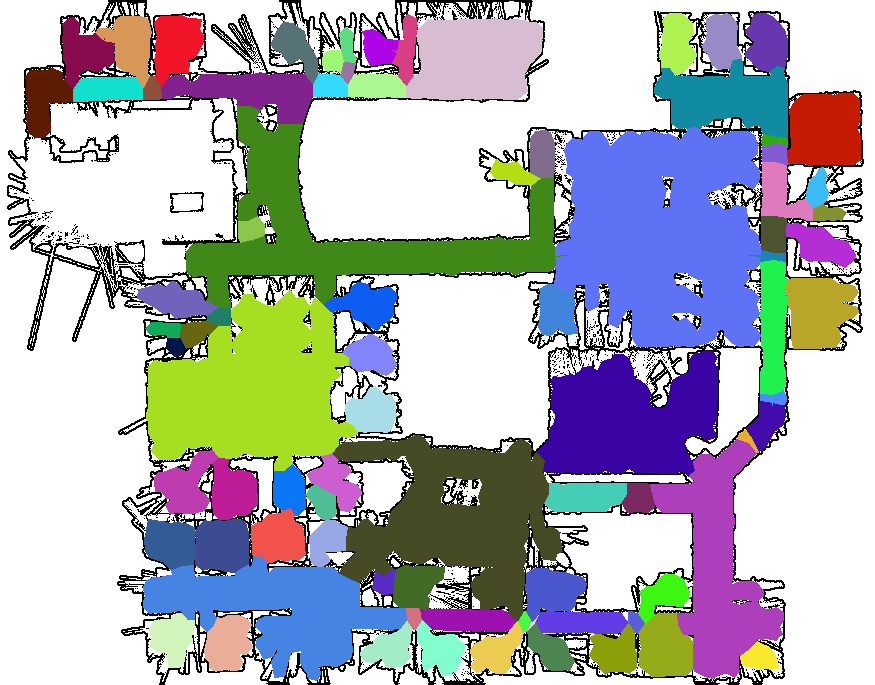}}
		\subfigure[Ours(strategy)]{ \includegraphics[width=0.15\linewidth]{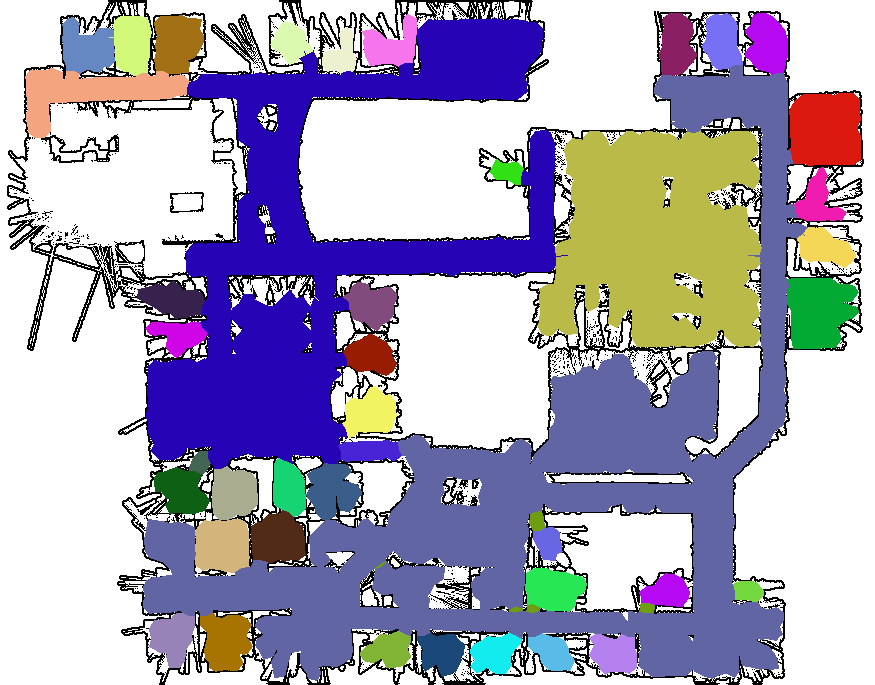}}
		\subfigure[Ground truth]{	\includegraphics[width=0.15\linewidth]{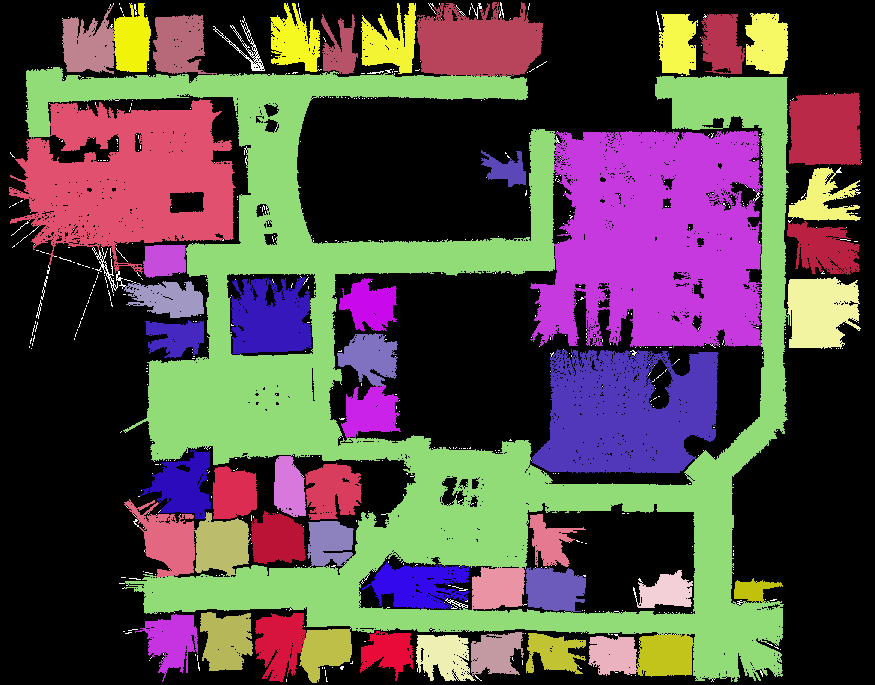}}
		\caption{Compare the segmentation results from three different methods. The input maps shown above are in this order: 1. Freiburg79\_scan\_furnitures\_trashbins; 2. lab\_c; 3. lab\_ipa\_furnitures; 4. Freiburg101\_scan\_furnitures\_trashbins; 5. lab\_d; 
		6. freiburg\_building52; 7. freiburg\_building101; 8. lab\_b; 9. lab\_e. I color the segmentation of MAORIS and the ground truths to make it easier to see. Although the different colors at the background, the evaluation algorithm didn't count the background. \label{fig:cmp_img}}
	\end{figure*}

	Defining what the ground truth segmentation of an environment is sometimes quite subjective and thus the MCC scores don't always reflect a final answer. Thus we also show the segmentation comparisons for some maps in Fig. \ref{fig:cmp_img} and have an analysis. 
	The MCC comparison results are plotted as bars and shown in Fig. \ref{fig:bars}. We found that in most cases our method performs better than the other two.
	$F79_{sft}$, $lab_c$, $lab_{ipasft}$ and $lab_{csft}$ are complete maps without complex furniture or big rooms, which hardly happen in practice. For this case the other two methods performed better than ours. 
	We got lower score for $lab_c$, $lab_{ipasft}$ and $lab_{csft}$ with fixed $W$ because of the very wide doors. Deciding the $\alpha$ with the strategy can improve the results. 
	Although $F101_s$, $lab_{d}$, $lab_{dsf}$ and $sistD$ are also complete maps without complex furniture however include big open space. 
	Thanks to the alpha shape room detection, our method performs much better in this case. 
	For the maps with furniture and noise in the middle, such as map $f52$, $f101$ and $lab_b$, especially map $lab_b$ contains contains a large number of clutter. Even though the pre-processing helps a lot, all of the three methods didn't  provide good results for $lab_b$. 
	Where the lower score of our method with strategy for $f52$ was mainly caused by the subjective ground truth. 
	Map $lab_b$ and $lab_e$ are incomplete scan maps, which causes under-segmentation and over-segmentation at the same time. But our algorithm still performed better, due to the reason that it detects the boundary of the map so that it can indicate the incomplete rooms as areas that are split to the unknown region.
	
	The mean scores of the MCCs for Voronoi-graph based segmentation by Bormann et al. \cite{bormann2016room}, the MAORIS method \cite{Malcolm2017MAORISICRA} and our method with fixed $W$ and $W$ decided by the strategy are 0.42, 0.47, 0.61 and 0.67, respectively. 
	Thus our method performed better than theirs, especially in cases with furniture, big open space or incomplete mapping. 

\subsection{Maze Experiment}
\begin{figure}[t]
	\subfigure{
		\label{fig:maze_ipa}	\includegraphics[width=0.31\linewidth]{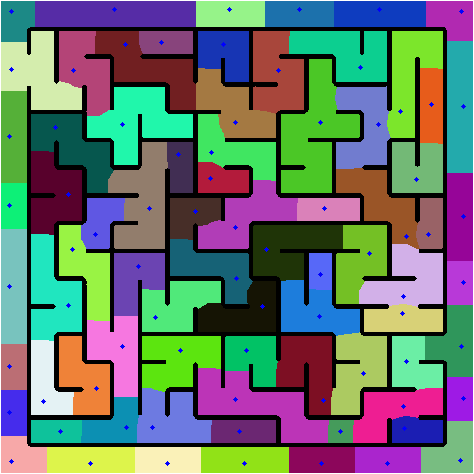}}
	\subfigure{
		\label{fig:maze_maoris}	\includegraphics[width=0.31\linewidth]{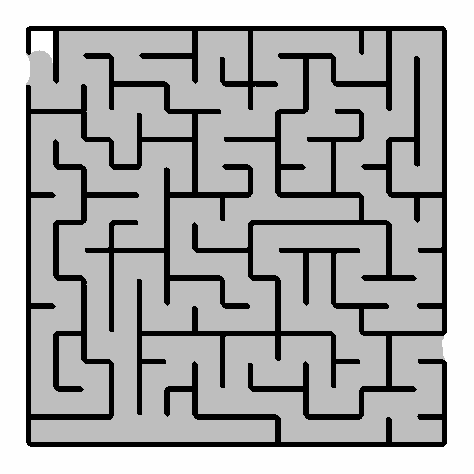}}
	\subfigure{
		\label{fig:maze_our}	\includegraphics[width=0.31\linewidth]{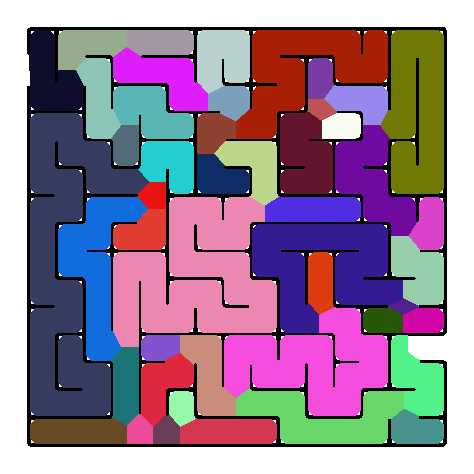}}
	\subfigure[Bormann's]{
		\label{fig:maze_ipa1}	\includegraphics[width=0.31\linewidth]{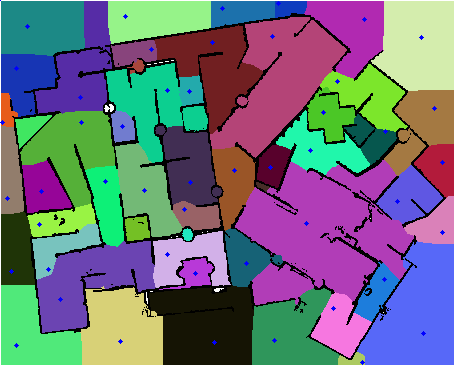}}
	\subfigure[MAORIS]{
		\label{fig:maze_maoris1}	\includegraphics[width=0.31\linewidth]{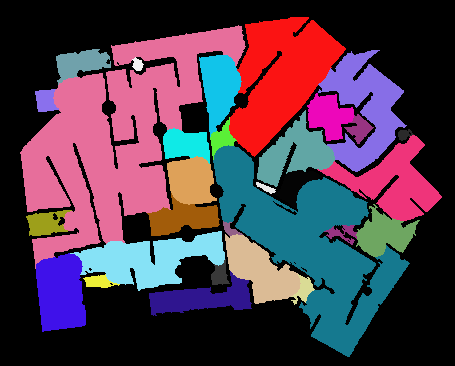}}
	\subfigure[Ours]{
		\label{fig:maze_our1}	\includegraphics[width=0.31\linewidth]{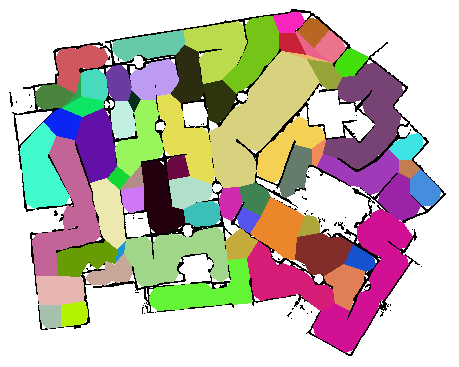}}
	\caption{Compare the segmentation methods on mazes. \label{fig:maze}}
\end{figure}

    Above we made that claim that our method extracts the underlying topology better than other methods, because it is directly based on the topology of the VD. Fig. \ref{fig:maze} depicts two example environments which highlight this.
    
    In the first environment we use an artificial maze as an extreme example. 
    The $\alpha$-shape value is set higher than the corridor width, so the room detection doesn't contribute here. 
    It can be seen that our algorithm captures the topology nicely, having areas from dead-ends to junctions and in-between junctions. The other two algorithms fail to generate meaningful areas. 
    
    The second environment is from RoboCup Rescue. We can see clear under-segmentation in MAORIS and over-segmentation in Bormann's. Ours isn't perfect, either, but still the best.
    
    

	\section{CONCLUSIONS}
	\label{sec:conclusions}
	
	This paper presents the Area Graph and the algorithm to automatically generate it from 2D grid maps. It is based on the Voronoi Diagram and a room segmentation using the $\alpha$-shape algorithm. 
	A comparison experiment to other segmentation methods shows that our method is more effective in segmentation for the complex maps. 
	
	The Area Graph can be used in many applications, for example path planning, localization \cite{Beeson2005Towards}, map merging\cite{Saeedi2014Group} or map evaluation\cite{schwertfeger2016map}. 
	
	As future work we plan to investigate hierarchical Area Graphs and graphs that store volumetric, topological 3D information.
	
	
	


	\addtolength{\textheight}{-11cm}   

	
	

	

	
	
	\bibliographystyle{IEEEtran}
	
	\bibliography{references}

\end{document}